\pdfoutput=1

\documentclass[11pt]{article}

\usepackage[final]{acl}

\usepackage{times}
\usepackage{latexsym}

\usepackage[T1]{fontenc}

\usepackage[utf8]{inputenc}

\usepackage{microtype}

\usepackage{inconsolata}

\usepackage{graphicx}

\usepackage{algorithm}
\usepackage{algorithmic}
\usepackage{bbding}
\usepackage{booktabs}
\usepackage{multirow}
\usepackage{amssymb}
\usepackage{makecell}
\usepackage{subfigure}
\usepackage{pifont}
\usepackage{newfloat}
\usepackage{listings}
\usepackage{xcolor}

\newcommand{\readoc}{\textsc{READoc}}

\title{\textsc{\readoc}: A Unified Benchmark for Realistic \\ Document Structured Extraction}

\author{
  Zichao Li${}^{1,2,}$\thanks{Equal contribution.},
  Aizier Abulaiti${}^{1,2,}$\footnotemark[1],
  \textbf{Yaojie Lu}${}^{1}$,
  Xuanang Chen${}^{1,}$\thanks{Corresponding author.}, \\
  \textbf{Jia Zheng}${}^{1,}$\footnotemark[2],
  \textbf{Hongyu Lin}${}^{1}$,
  \textbf{Xianpei Han}${}^{1}$,
  \textbf{Shanshan Jiang}${}^{3}$,
  \textbf{Bin Dong}${}^{3}$,
  \textbf{Le Sun}${}^{1}$
  \\
  ${}^{1}$Chinese Information Processing Laboratory, Institute of Software,
  Chinese Academy of Sciences \\
  ${}^{2}$University of Chinese Academy of Sciences \\
  ${}^{3}$Ricoh Software Research Center Beijing Co., Ltd \\
 {\tt \{lizichao2022,aizier2022,luyaojie,chenxuanang,zhengjia\}@iscas.ac.cn} \\
   {\tt \{hongyu,xianpei,sunle\}@iscas.ac.cn \{shanshan.jiang,bin.dong\}@cn.ricoh.com} \\
}

\begin{document}
\maketitle
\begin{abstract}
Document Structured Extraction (DSE) aims to extract structured content from raw documents. 
Despite the emergence of numerous DSE systems,
their unified evaluation remains inadequate, significantly hindering the field’s advancement.
This problem is largely attributed to existing benchmark paradigms, which exhibit fragmented and localized characteristics. 
To offer a thorough evaluation of DSE systems, we introduce a novel benchmark named \readoc, which defines DSE as a realistic task of converting unstructured PDFs into semantically rich Markdown. 
The \readoc~ dataset is derived from 3,576 diverse and real-world documents from arXiv, GitHub, and Zenodo. 
In addition, we develop a DSE Evaluation S$^3$uite comprising Standardization, Segmentation and Scoring modules, to conduct a unified evaluation of state-of-the-art DSE approaches.
By evaluating a range of pipeline tools, expert visual models, and general Vision-Language Models, we identify the gap between current work and the unified, realistic DSE objective for the first time.
We aspire that \readoc~will catalyze future research in DSE, fostering more comprehensive and practical solutions. 
\end{abstract}

\section{Introduction}

The wealth of knowledge preserved in documents is immeasurable.
Document Structured Extraction (DSE), which involves converting raw documents into machine-readable structured text \cite{tkaczyk2015cermine,zhong2019publaynet,shen2022vila,lo2023papermage}, is increasingly crucial in real-world scenarios. It facilitates building extensive knowledge bases \cite{wang2020cord}, constructing high-quality corpora \cite{jain2020scirex}, and plays a pivotal role in Retrieval-Augmented Generation (RAG) \cite{gao2023retrieval} applications for Large Language Models (LLMs) \cite{achiam2023gpt}.

\begin{figure}[t]
\centering
\includegraphics[width=0.45\textwidth]{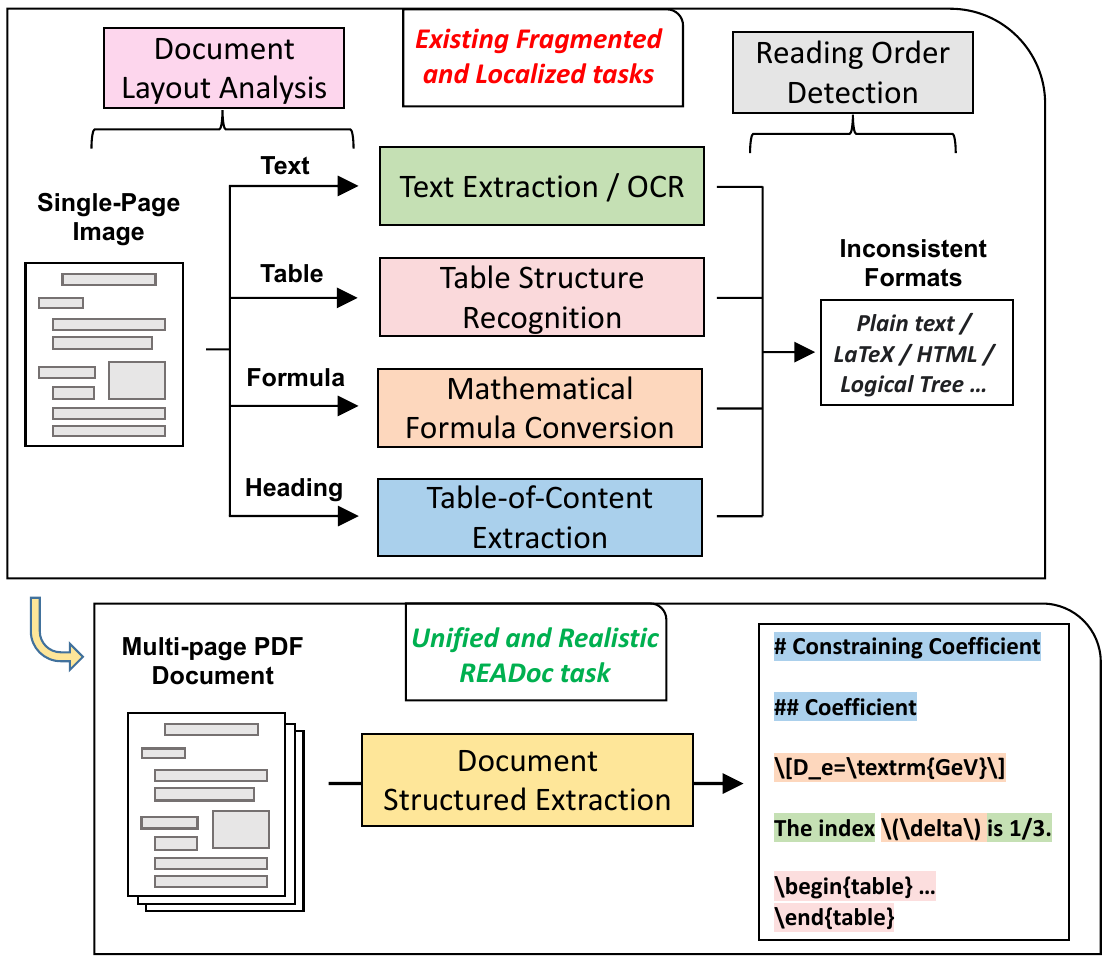} 
\caption{A comparison between fragmented and localized DSE task views and the \readoc~benchmark paradigm.}
\label{fig1}
\end{figure}

\begin{table*}[ht] \small
  \centering
    \setlength{\tabcolsep}{1mm}
    \scalebox{0.95}{
    \begin{tabular}{c|cccccccc}
    \toprule
    \multirow{2}[5]{*}{\textbf{Benchmark}} & \multicolumn{6}{c}{\textbf{Coverage}}         & \multicolumn{2}{c}{\textbf{Task Paradigm}} \\
\cmidrule{2-9}          & \makecell{\textit{Layout} \\ \textit{Analysis}} & \makecell{\textit{Character} \\ \textit{Recognition}} & \makecell{\textit{Table} \\ \textit{Recognition}} & \makecell{\textit{Formula} \\ \textit{Conversion}} & \makecell{\textit{ToC.} \\ \textit{Extraction}} & \makecell{\textit{Order} \\ \textit{Detection}} & \textsc{Input} & \textsc{Output} \\
    \midrule
    PubLayNet &  \Checkmark  &   \ding{55}    &   \ding{55}    &    \ding{55}   &    \ding{55}   &   \ding{55}    & \textit{Page image} & \textit{Layout blocks} \\
    DocBank &  \Checkmark   &   \ding{55}    &    \ding{55}   &  \ding{55}     &    \ding{55}   &   \ding{55}    & \textit{Page image \& text} & \textit{Layout blocks} \\
    Robust Reading &  \ding{55}     &  \Checkmark   &   \ding{55}    &  \ding{55}     &   \ding{55}    &   \ding{55}    & \textit{Text image} & \textit{Plain text} \\
    PubTabNet &   \ding{55}    &    \ding{55}   &  \Checkmark   &   \ding{55}    &  \ding{55}     &    \ding{55}   & \textit{Table image} & \textit{HTML table} \\
    TabLeX &   \ding{55}    &  \ding{55}   &  \Checkmark   &   \ding{55}    &   \ding{55}    &   \ding{55}    & \textit{Table image} & \textit{\LaTeX~table} \\
    Im2Latex-100K &   \ding{55}    &   \ding{55}    &   \ding{55}    &  \Checkmark   &   \ding{55}    &    \ding{55}   & \textit{Formula image} & \textit{\LaTeX~formula} \\
    ReadingBank &  \ding{55}     &    \ding{55}   &   \ding{55}    &   \ding{55}    &   \ding{55}    &  \Checkmark   & \textit{Unsorted tokens} & \textit{Sorted tokens} \\
    HRDoc &  \Checkmark   &   \ding{55}    &   \ding{55}    &   \ding{55}    &  \Checkmark   &    \ding{55}   & \textit{Doc images \& text} & \textit{Logical tree} \\ 
    \midrule
    \textbf{\readoc} &  \Checkmark   &  \Checkmark   &   \Checkmark  &   \Checkmark  &  \Checkmark   &  \Checkmark   & \textit{Realistic PDF doc} & \textit{Markdown text} \\
    \bottomrule
    \end{tabular}
    }
      \caption{A comparison between \readoc~and existing DSE benchmarks, including PubLayNet~\cite{zhong2019publaynet}, DocBank~\cite{li2020docbank}, Robust Reading~\cite{karatzas2015icdar}, PubTabNet~\cite{zhong2020image}, TabLeX~\cite{desai2021tablex}, Im2Latex-100K~\cite{deng2017image}, ReadingBank~\cite{wang2021layoutreader}, and HRDoc~\cite{ma2023hrdoc}.}
  \label{tab:compare}%
\end{table*}%

Recent progress in Document AI  \cite{appalaraju2021docformer,huang2022layoutlmv3,ye2023ureader,hu2024mplug} has led to the creation of numerous DSE systems \cite{pix2text,marker,blecher2023nougat}.
However, the absence of unified evaluation in real-world scenarios has left uncertainty about their performance levels and hindered their further development.
This issue is largely due to the limitations of prevailing benchmark paradigms, which exhibit \textbf{{fragmented}} and \textbf{{unrealistic}} characteristics.
Firstly, as depicted in Figure \ref{fig1}, \textbf{{existing benchmarks typically fragment DSE into distinct subtasks}}, including document layout analysis \cite{zhong2019publaynet}, optical character recognition \cite{karatzas2015icdar}, table-of-contents extraction \cite{hu2022multimodal}, reading order detection \cite{wang2021layoutreader}, table recognition \cite{smock2022pubtables} and formula conversion \cite{deng2017image}. 
Due to their narrow focus, diverse data sources, and inconsistent formats, existing benchmarks lack a unified framework to comprehensively evaluate DSE systems. Additionally, \textbf{{current research often unrealistically targets localized regions}}, such as layout blocks or tables within a single page. This approach overlooks the complexity of real-world documents, which typically span multiple pages with hierarchical headings and require long-range dependencies to construct a global structure. Benchmarks focusing on isolated pages or blocks fail to provide realistic evaluations.

To address these issues, we introduce {\readoc}, a unified benchmark designed to quantify the gap between existing work and the goal of \textsc{REA}listic \textsc{Doc}ument Structured Extraction. 
\readoc~formulates DSE as an end-to-end task, converting multi-page PDFs into structured Markdown. We automatically construct 3,576 PDF-Markdown pairs from arXiv, GitHub, and Zenodo, covering diverse types, years, and topics to reflect real-world complexity. 
Additionally, we develop a DSE evaluation S$^3$uite with three modules: Standardization, Segmentation, and Scoring, enabling unified evaluation of diverse DSE systems, including pipeline tools \cite{marker}, expert models \cite{blecher2023nougat}, and Vision-Language Models \cite{achiam2023gpt}.

Our contributions are three-fold:
1) \readoc~is the first benchmark to frame DSE as a PDF-to-Markdown paradigm, which is realistic, end-to-end, and incorporates diverse data.
2) An evaluation S$^3$uite is proposed to support the unified assessment of various DSE systems and to quantify multiple capabilities required for DSE.
3) We present the gap between current research and realistic DSE, emphasizing the importance of exploring new modeling paradigms.
The code and data are publicly available at \url{https://github.com/icip-cas/READoc}.

\section{Related Work}

\subsection{Task Views and Benchmarks}



DSE is a crucial task, yet existing benchmarks focus on discrete subtasks:
{document layout analysis} \cite{zhong2019publaynet,li2020docbank} identifies layout blocks;
{optical character recognition} \cite{karatzas2015icdar} extracts text from images;
{table structure recognition} \cite{zhong2020image} transforms tables into structured formats;
{mathematical formula conversion} \cite{deng2017image} converts formulas into semantic formats;
{table-of-contents (ToC) extraction} \cite{ma2023hrdoc} constructs hierarchical heading trees;
{reading order detection} \cite{wang2021layoutreader} sorts page elements by reading order.
We summarize relevant benchmarks in Table \ref{tab:compare}. However, their heterogeneity complicates unified DSE evaluation.

Recent research conceptualizes DSE as a single-page image-to-markup task \cite{blecher2023nougat,lee2023pix2struct}, and targeted benchmarks such as OmniDocBench \cite{ouyang2024omnidocbench} have emerged. Although OmniDocBench demonstrates commendable richness in evaluation and diversity of data, its single-page approach falls short in handling multi-page or lengthy documents often encountered in real-world scenarios.

\subsection{Methods for Document Structured Extraction}

Due to the intricacies of textual, graphical, and layout information within documents \cite{xu2020layoutlm}, a universally accepted method for DSE has yet to emerge. 
A common simplistic strategy involves leveraging external parsing engines \cite{PyMuPDF4LLM} to extract text and metadata from digital-born PDFs.
With the rise of deep learning techniques, Numerous systems \cite{li2022pp,marker,2024mineru} have integrated a series of submodels into a pipeline, with each submodel dedicated to a specific subtask of DSE. 
Recent advancements have shifted towards end-to-end DSE methodologies, with some works leveraging Transformer \cite{vaswani2017attention} expert models to convert document page images directly into structured formats such as HTML \cite{lee2023pix2struct} or Markdown \cite{blecher2023nougat}.

Recently, large Vision-Language Models have garnered widespread attention \cite{achiam2023gpt,liu2024visual}, 
with document understanding \cite{feng2023docpedia,hu2024mplug} being a key focus of their capabilities. 
Efforts to enhance VLMs' document understanding capability have focused on methods like tailored training tasks \cite{ye2023ureader} and resolution adaptation \cite{li2024monkey}, leading to notable improvements. VLMs obtain impressive results on various DSE subtasks, such as OCR \cite{liu2024textmonkey}, table recognition \cite{zhao2024tabpedia}, and formula conversion \cite{xia2024docgenome}. Additionally, some research has explored converting page images into structured text using VLMs \cite{lv2023kosmos,wei2024small,liu2024focus}.
However, the lack of a unified benchmark leaves uncertainty about the gap between VLMs' current capabilities and realistic DSE needs.

\section{Task Definition}


We establish a realistic task paradigm for end-to-end DSE, using raw PDF documents as input due to their prevalence and unstructured nature, which poses challenges with dispersed content and multimodal information. On the other hand, we employ Markdown as the output format, leveraging its lightweight markup to represent structural elements like headings and lists.
We adopt a variant of Markdown \cite{blecher2023nougat} that supports \LaTeX~syntax for tables and formulas. Markdown, as the target format, can be chunked, indexed as flat text, or directly ingested by LLMs.

In summary, \readoc~uniformly defines DSE as a task that takes a complete PDF file as input and generates structured text in Markdown format, which is well-defined, practical, and challenging for DSE systems.
Examples of task inputs and outputs are provided in Appendix \ref{task_example}.

\section{Benchmark Construction}

\readoc~is a unified DSE benchmark derived from real-world documents. We select heterogeneous documents from arXiv preprints\footnote{\url{https://arxiv.org/}}, GitHub READMEs\footnote{\url{https://github.com/}} and Zenodo's Open Research Repository\footnote{\url{https://zenodo.org/}}, which are then automatically processed to construct PDF-Markdown pairs.
\readoc~consists of 3,576 documents: 1,009 in the \readoc-arXiv subset, 1,224 in the \readoc-GitHub subset, and 1343 in the \readoc-Zenodo subset. Each subset offers unique challenges: \readoc-arXiv features complex academic structures such as formulas and tables, with diverse multi-column layout templates. However, its heading styles are simple, often following easily recognizable patterns like "1.1 Introduction." In contrast, \readoc-GitHub includes only basic elements like paragraphs and headings, and presents a uniform single-column layout style. However, building its ToC structure is challenging due to varied and often unmarked heading styles. 
\readoc-Zenodo contains longer documents (many exceeding 30 pages) and diverse types, such as posters, reports, theses, and books, challenging layout analysis. Additionally, its 27 languages further increase text processing and semantic understanding complexity.
Each subset exhibits significant diversity in types, topics, eras, and so on, establishing \readoc~as a robust benchmark. 
We describe more construction details in Appendix \ref{dataset_detail}.


\begin{figure*}[ht]
\centering
\subfigure[]{
\centering
\includegraphics[height=1.2in, keepaspectratio]{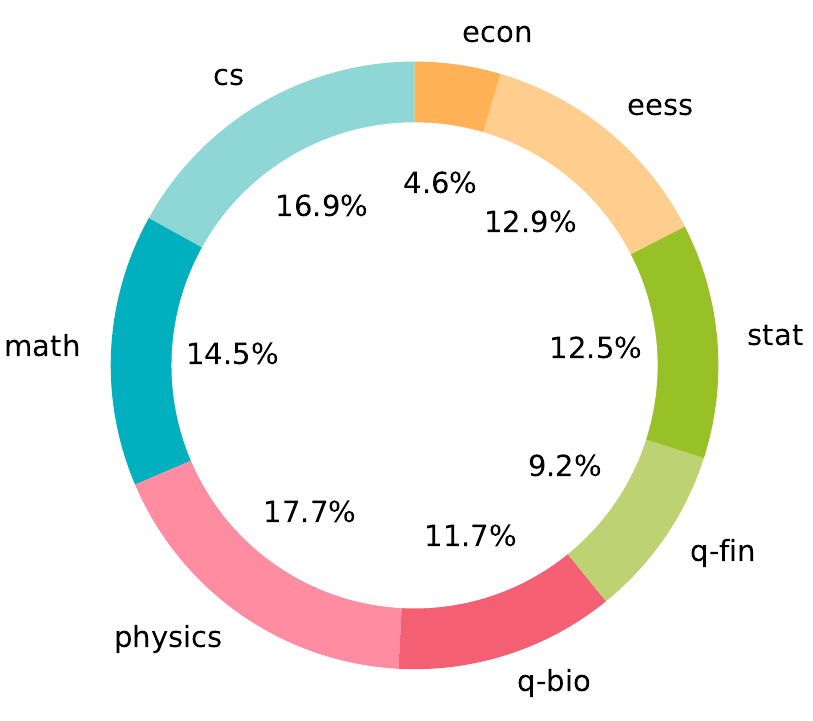}
}%
\subfigure[]{
\raggedright
\includegraphics[height=1.2in, keepaspectratio]{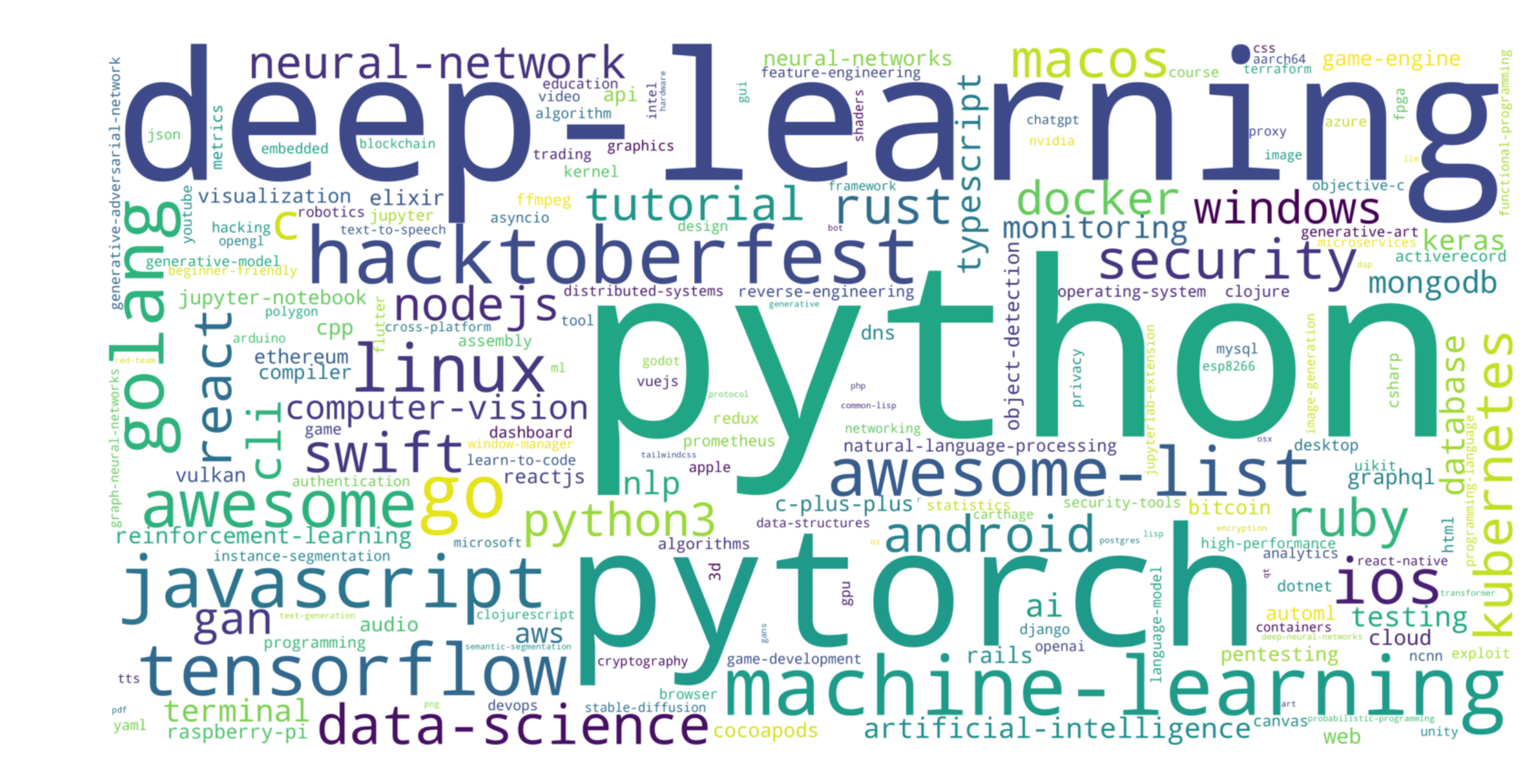}
}%
\subfigure[]{
\raggedright
\includegraphics[height=1.2in, keepaspectratio]{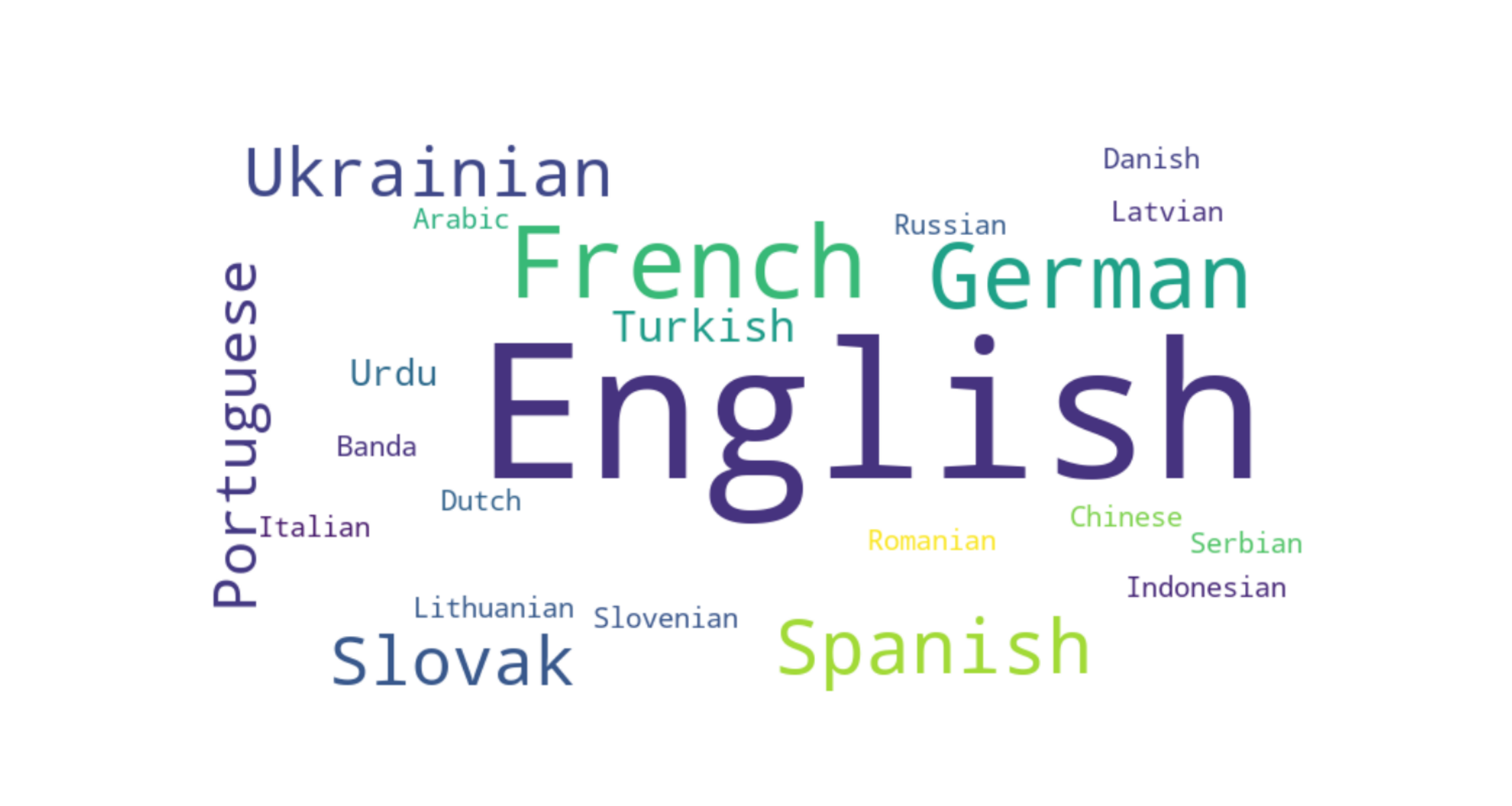}  
}%
\caption{Visualization of data distribution in \readoc. (a) Document disciplines of \readoc-arXiv. (b) Document topics of \readoc-GitHub. (c) Language distribution of \readoc-Zenodo. 
}
\label{fig.diversity}
\end{figure*}

\subsection{Document Collection and Processing}


\paragraph{\readoc-arXiv.}
For the arXiv preprints, the collection process involves using type keywords such as ``Conference" and ``Journal" to select papers and filtering for English language. Preprints without \LaTeX~files or with unclear main \LaTeX~files are excluded. The selected documents are first converted from \LaTeX~to HTML using LaTeXML\footnote{\url{https://github.com/brucemiller/LaTeXML}}, followed by a conversion from HTML to Markdown using a modified version of the Nougat \cite{blecher2023nougat} process.
Only documents that complete this process without any errors and maintain correct table syntax after conversion are included.

\paragraph{\readoc-GitHub.}
The GitHub README files are originally in Markdown format, and we collect and filter them based on specific criteria: they must have obtained than 500 stars, be written in English, exclude HTML syntax, etc. To maintain the simplicity of this subset, we exclude files that contain tables and formulas. After initial preprocessing, the Markdown files are converted to PDFs using Pandoc\footnote{\url{https://github.com/jgm/pandoc}} as the conversion engine and Eisvogel\footnote{\url{https://github.com/Wandmalfarbe/pandoc-latex-template}} as the template. Only documents that successfully complete this entire workflow without execution errors or warnings are retained.

\paragraph{\readoc-Zenodo.}
For the Zenodo dataset, we collect 910 DOCX and 433 HTML files from Zenodo. DOCX files are converted to Markdown using Microsoft's Markitdown~\cite{MarkItDown} tool, then to PDF via python-docx\footnote{\url{https://pypi.org/project/python-docx/}}. HTML files are converted to Markdown using Pandoc\footnote{\url{https://github.com/jgm/pandoc}}, then to PDF via Chromium. We exclude conversion failures and overly long/short files, retaining only structurally rich documents.




\begin{table}[t] \small
    \centering
    \setlength{\tabcolsep}{1mm}
    \scalebox{0.95}{
    \begin{tabular}{lccc}
    \toprule
       \makebox[0.15\textwidth][l]{\textbf{Statistics}}  & \textbf{arXiv} & \textbf{GitHub} & \textbf{Zenodo}  \\ \midrule
       Documents & 1,009 & 1,224 & 1,343 \\
       Avg. Pages & 11.67 & 6.54 & 14.93 \\
       Avg. Depth & 3.10 & 3.11 & 2.66 \\
       Avg. Length & 10,209.50 & 1,978.10 & 8255.85 \\ 
        Year Span & 1996 - 2024  & 2008 - 2024 & 2014 - 2024 \\
        Types / Disciplines & 6 / 8 & - & - \\
        Topics & - & 2,805 & - \\
        Language count & 1 & 1 & 27 \\
    \bottomrule
    \end{tabular} }
    \caption{The data statistics of \readoc.}
    \label{tab:statistics}
\end{table}

\begin{figure*}[ht]
\centering
\includegraphics[width=0.83\textwidth]{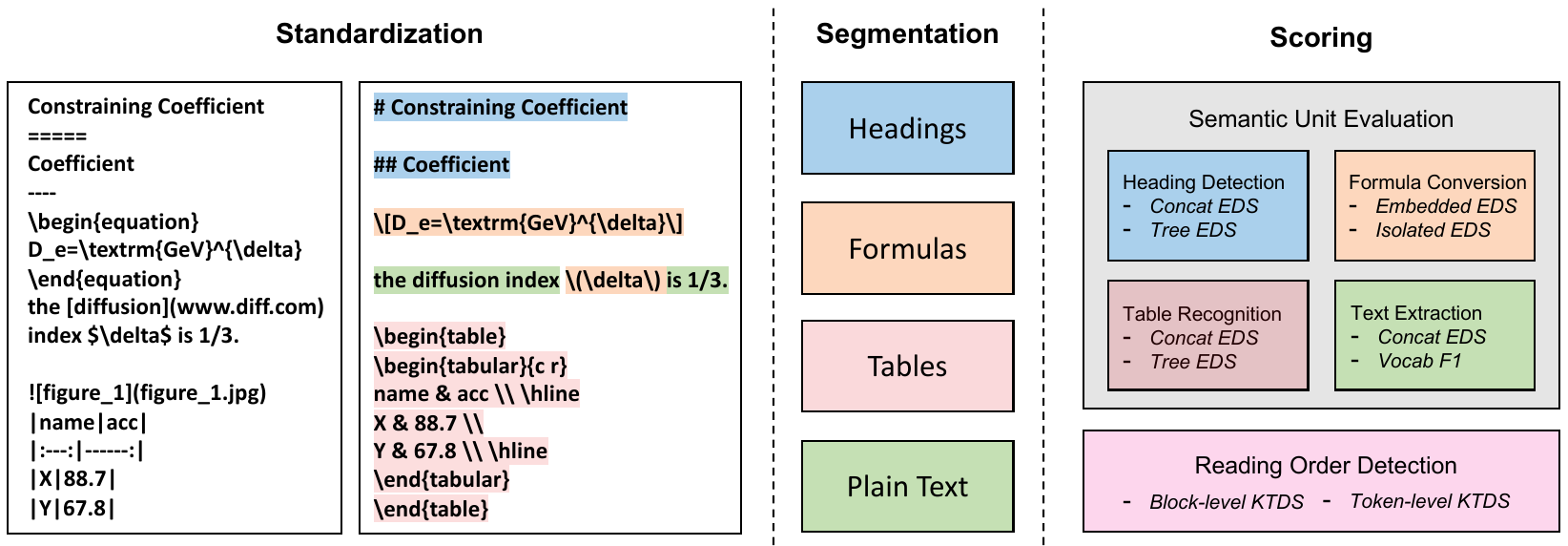} 
\caption{The three modules of the \readoc~Evaluation S$^3$uite: Standardization, Segmentation and Scoring.}
\label{fig2}
\end{figure*}

\subsection{Dataset Statistics}

We present the basic statistics of \readoc~in Table \ref{tab:statistics}, showcase the diversity of \readoc~in Figure \ref{fig.diversity}, and provide additional statistics in Appendix \ref{dataset_detail}. 
Overall, our benchmark is divided into three subsets. 
\readoc-arXiv consists of 1,009 documents, with an average of 11.67 pages, 10,209.50 tokens, and 3.10 heading levels. These documents cover a timeline from 1996 to 2024, ensuring ample diversity across 6 types and 8 disciplines.
\readoc-GitHub comprises 1,224 documents, with an average of 6.54 pages, 1,978.11 tokens, and 3.11 heading levels. These documents are sourced from projects spanning the years 2008 to 2024, encompassing a rich tapestry of 2,805 topics.
\readoc-Zenodo contains 1,343 documents, with an average of 14.93 pages, 8,255.85 tokens, and 2.66 heading levels. These documents span from 2014 to 2024 and include 27 languages, e.g., English and French.

\section{Evaluation S$^3$uite}

Considering the potential confusion that various models may have with our Markdown syntax and the multifaceted nature of the capabilities required for DSE tasks, we propose an Evaluation S$^3$uite, consisting of three sequential modules: Standardization, Segmentation, and Scoring,
as shown in Figure~\ref{fig2}. 
The S$^3$uite ensures that \readoc~can automatically perform a unified evaluation of the end-to-end DSE task, yielding reliable and effective assessment results.
More implementation details are in Appendix \ref{suite_detail}.

\subsection{Standardization}

The first module of the S$^3$uite standardizes the output Markdown text to align with the ground truth Markdown format. This alignment is essential for mitigating the impact on evaluation accuracy caused by variations in the format and syntax of texts generated by different DSE systems.
The standardization includes aligning the boundaries of formulas, such as \textit{\$\$}, \textit{\textbackslash begin\{equation\}}, and \textit{\textbackslash [}; unifying different Markdown heading styles; aligning Markdown tables with \LaTeX~table formats for evaluation consistency; removing images and eliminating the link syntax, etc.
This module ensures that the focus remains on the core DSE capabilities, rather than being clouded by formatting disunity.

\subsection{Segmentation}

The second module of the S$^3$uite divides the standardized Markdown text into distinct semantic units. To facilitate \readoc~in highlighting various specialized DSE capabilities of the models within a single document, we divide both the output and the ground truth Markdown text into four units: \texttt{Headings} of different levels, \texttt{Formulas} in both embedded and isolated forms, \texttt{Tables}, and residual \texttt{Plain Text} encompassing basic text and simple formatting such as bold, italic, and lists. 

\begin{table*}[ht] \small
  \centering
  \setlength{\tabcolsep}{1mm}
  \scalebox{0.95}{
    \begin{tabular}{c|cccccccc|cc|c}
    \toprule
    \multirow{3}[2]{*}{\textbf{Methods}} & \multicolumn{8}{c|}{\textbf{Semantic Unit Evaluation}}             & \multicolumn{2}{c|}{\multirow{2}[2]{*}{\makecell{\textbf{Reading} \\ \textbf{Order}}}} & \multirow{3}[2]{*}{\textbf{Average}} \\ \cmidrule{2-9}
          & \multicolumn{2}{c}{\textbf{Text}} & \multicolumn{2}{c}{\textbf{Heading}} & \multicolumn{2}{c}{\textbf{Formula}} & \multicolumn{2}{c|}{\textbf{Table}} &  &  \\
          & \textit{Concat} & \textit{Vocab} & \textit{Concat} & \textit{Tree} & \textit{Embed} & \textit{Isolate} & \textit{Concat} & \textit{Tree} & \textit{Block} & \textit{Token} &  \\
    \midrule
    \multicolumn{12}{c}{\textbf{Baselines}} \\
    \midrule
    PyMuPDF4LLM & 66.66  & 74.27  & \textbf{27.86}  & \textbf{20.77}  & 0.07  & \textbf{0.02}  & \textbf{23.27}  & \textbf{15.83}  & 87.70  & 89.09  & \textbf{40.55}  \\
    Tesseract OCR & \textbf{78.85}  & 76.51  & 1.26  & 0.30  & \textbf{0.12}  & 0.00  & 0.00  & 0.00  & 96.70  & \textbf{97.59}  & 35.13  \\
    MarkItDown & 73.88  & \textbf{79.64}  & 2.90  & 0.93  & 0.10  & 0.00  & 0.34  & 0.08  & \textbf{97.13}  & 97.08  & 35.21  \\
    \midrule
    \multicolumn{12}{c}{\textbf{Pipeline Tools}} \\
    \midrule
    MinerU & \textbf{88.32}  & \textbf{91.22}  & 67.06  & \textbf{41.97}  & \textbf{62.77}  & \textbf{70.76}  & 59.34  & 52.85  & \textbf{98.52}  & \textbf{97.90}  & \textbf{73.07}  \\
    Pix2Text & 85.85  & 83.72  & 63.23  & 34.53  & 43.18  & 37.45  & 54.08  & 47.35  & 97.68  & 96.78  & 64.39  \\
    Marker & 79.11  & 82.71  & 63.60 & 39.39  & 3.47  & 48.74  & \textbf{64.61}  & \textbf{72.36}  & 98.04  & 97.74  & 64.98  \\
    Docling & 79.73 & 85.39 & \textbf{68.74} & 38.33 & 0.23 & 0.0 & 54.09 & 66.56 & 98.05 & 97.18 & 58.83 \\
    \midrule
    \multicolumn{12}{c}{\textbf{Expert Visual Models}} \\
    \midrule
    Nougat-small & 87.35  & 92.00  & 86.40  & 87.88  & \textbf{76.52} & 79.39  & \textbf{55.63} & \textbf{52.35} & 97.97  & 98.36  & 81.38  \\
    Nougat-base & \textbf{88.03} & \textbf{92.29} & \textbf{86.60} & \textbf{88.50} & 76.19  & \textbf{79.47} & 54.40  & 52.30  & \textbf{97.98} & \textbf{98.41} & \textbf{81.42} \\
    GOT-OCR 2.0 & 84.47 & 86.24 & 66.69 & 57.68 & 53.48 & 56.23 & 50.40 & 34.50 & 97.73 & 97.50 & 68.49 \\
    \midrule
    \multicolumn{12}{c}{\textbf{Vision-Language Models}} \\
    \midrule
    DeepSeek-VL-7B-Chat & 31.89  & 39.96  & 23.66  & 12.53  & 17.01  & 16.94  & 22.96  & 16.47  & 88.76  & 66.75  & 33.69  \\
    MiniCPM-Llama3-V2.5 & 58.91  & 70.87  & 26.33  & 7.68  & 16.70  & 17.90  & 27.89  & 24.91  & {95.26}  & {93.02}  & 43.95  \\
    LLaVa-1.6-Vicuna-13B & 27.51  & 37.09  & 8.92  & 6.27  & 17.80  & 11.68  & 23.78  & 16.23  & 76.63  & 51.68  & 27.76  \\
    InternVL-Chat-V1.5 & 53.06  & 68.44  & 25.03  & 13.57  & 33.13  & 24.37  & 40.44  & 34.35  & 94.61  & 91.31  & 47.83  \\
    GPT-4o-mini & \textbf{79.44}  & \textbf{84.37}  & \textbf{31.77}  & \textbf{18.65}  & \textbf{42.23}  & \textbf{41.67}  & \textbf{47.81}  & \textbf{39.85}  & \textbf{97.69}  & \textbf{96.35}  & \textbf{57.98}  \\
    \bottomrule
    \end{tabular} }
      \caption{Evaluation of various Document Structured Extraction systems on \readoc-arXiv.}
  \label{tab:arxiv_result}%
\end{table*}%

\subsection{Scoring} \label{score}

The scoring module comprises two submodules:

The \textbf{Semantic Unit Evaluation} submodule implicitly assesses the models' layout analysis ability of identifying semantic units and explicitly measures four specialized capabilities:
1) \textbf{Text Extraction} refers to extracting plain text through PDF bytecode parsing or visual methods (e.g., OCR tools, VLMs). We measure edit distance similarity (EDS) after concatenating plain text and the F1 score of the plain text vocabulary.
2) \textbf{Heading Detection} involves detecting headings and constructing a hierarchical ToC tree. We measure EDS after concatenating all headings and construct ToC trees to calculate tree edit distance similarity (TEDS).
3) \textbf{Formula Conversion} involves transforming mathematical formulas into \LaTeX~format. We measure EDS by concatenating all embedded formulas, as well as all isolated formulas.
4) \textbf{Table Recognition} involves identifying tables' structures. We evaluate EDS after concatenating all tables. Then, we convert tables into structural trees, use a maximum bipartite matching algorithm to find the optimal mapping between tables and calculate TEDS for the matched tables.

The \textbf{Reading Order Detection} submodule determines whether the model extracts document elements in the correct order. We first segment each document into blocks based on semantic unit boundaries and line breaks, create two ordered lists from these blocks, and calculate Kendall's Tau Distance Similarity (KTDS) between the lists. Additionally, we divide each document into sequential tokens, construct lists based on the positions of co-occurring tokens at their first appearance, and compute KTDS between the token-level lists.


\section{Experiments}

\begin{figure*}[]
 \scriptsize
 \begin{minipage}{0.57\textwidth} \small
  \centering
    \scriptsize
    \setlength{\tabcolsep}{1mm}
    \resizebox{\textwidth}{!}{
        \begin{tabular}{c|cccc|cc|c}
        \toprule
        \multirow{3}[2]{*}{\textbf{Methods}} & \multicolumn{4}{c|}{\textbf{Semantic Unit Evaluation}} & \multicolumn{2}{c|}{\multirow{2}[2]{*}{\makecell{\textbf{Reading} \\ \textbf{Order}}}} & \multirow{3}[2]{*}{\textbf{Avg.}} \\
    \cmidrule{2-5}          & \multicolumn{2}{c}{\textbf{Text}} & \multicolumn{2}{c|}{\textbf{Heading}} & \multicolumn{2}{c|}{} &  \\
              & \textit{Concat} & \textit{Vocab} & \textit{Concat} & \textit{Tree} & \textit{Block} & \textit{Token} &  \\
        \midrule
        \multicolumn{8}{c}{\textbf{Baselines}} \\
        \midrule
        PyMuPDF4LLM & 85.21  & 77.27  & \textbf{12.13}  & \textbf{11.05}  & 98.61  & 98.43  & \multicolumn{1}{c}{\textbf{63.78}} \\
        Tesseract OCR & 82.06  & 82.32  & 6.65  & 4.25  & 98.48  & 99.01 & \multicolumn{1}{c}{62.13} \\
        MarkItDown & \textbf{85.23}  & \textbf{85.69}  & 2.07  & 0.64  & \textbf{99.4}  & \textbf{99.42} & \multicolumn{1}{c}{62.07} \\
        \midrule
        \multicolumn{8}{c}{\textbf{Pipeline Tools}} \\
        \midrule
        MinerU & 84.46  & 84.78  & 67.24 & \textbf{47.15} & \textbf{99.51} & \textbf{99.18}  & \multicolumn{1}{c}{80.39} \\
        Pix2Text & 78.99  & 78.51  & 60.53  & 39.42  & 97.94  & 97.38  & \multicolumn{1}{c}{75.46} \\
        Marker & \textbf{89.50} & \textbf{88.11} & \textbf{72.81}  & 37.51  & 99.03  & 99.13  & \multicolumn{1}{c}{\textbf{81.02}} \\
        Docling & 44.53 & 50.73 & 40.17 & 22.8 & 70.77 & 70.19 & \multicolumn{1}{c}{49.87} \\
        \midrule
        \multicolumn{8}{c}{\textbf{Expert Visual Models}} \\
        \midrule
        Nougat-small & 76.04  & 77.11  & \textbf{62.81}  & 38.73  & 98.33  & 96.12  & \multicolumn{1}{c}{74.86} \\
        Nougat-base & 75.26  & 76.79  & 62.39  & 37.01  & 97.65  & 95.62  & \multicolumn{1}{c}{74.12} \\
        GOT-OCR 2.0 & \textbf{78.18} & \textbf{84.53} & 62.08 & \textbf{50.06} & \textbf{98.40} & \textbf{98.60} & \multicolumn{1}{c}{\textbf{78.64}} \\
        \midrule
        \multicolumn{8}{c}{\textbf{Vision-Language Models}} \\
        \midrule
        DeepSeek-VL-7B-Chat & 34.85  & 41.97  & 38.69  & 20.15  & 96.47  & 84.55  & \multicolumn{1}{c}{52.78} \\
        MiniCPM-Llama3-V2.5 & 71.61  & 78.88  & 39.66  & 22.99  & 97.91  & 97.97  & \multicolumn{1}{c}{68.17} \\
        LLaVa-1.6-Vicuna-13B & 37.03  & 52.88  & 27.49  & 16.13  & 95.65  & 90.45  & \multicolumn{1}{c}{53.27} \\
        InternVL-Chat-V1.5 & 72.59  & 78.27  & 56.85  & 32.70  & 98.39  & 97.69  & \multicolumn{1}{c}{72.75} \\
        GPT-4o-mini & \textbf{85.06}  & \textbf{89.41}  & \textbf{62.65}  & \textbf{42.04}  & \textbf{98.81}  & \textbf{99.05}  & \textbf{79.50}  \\
        \bottomrule
        \end{tabular}%
    }
    
  \captionof{table}{Evaluation of various DSE systems on \readoc-GitHub.}
  \label{tab:github_result}%
 \end{minipage}
 \begin{minipage}{0.38\textwidth}
 \centering
\includegraphics[width=0.82\linewidth]{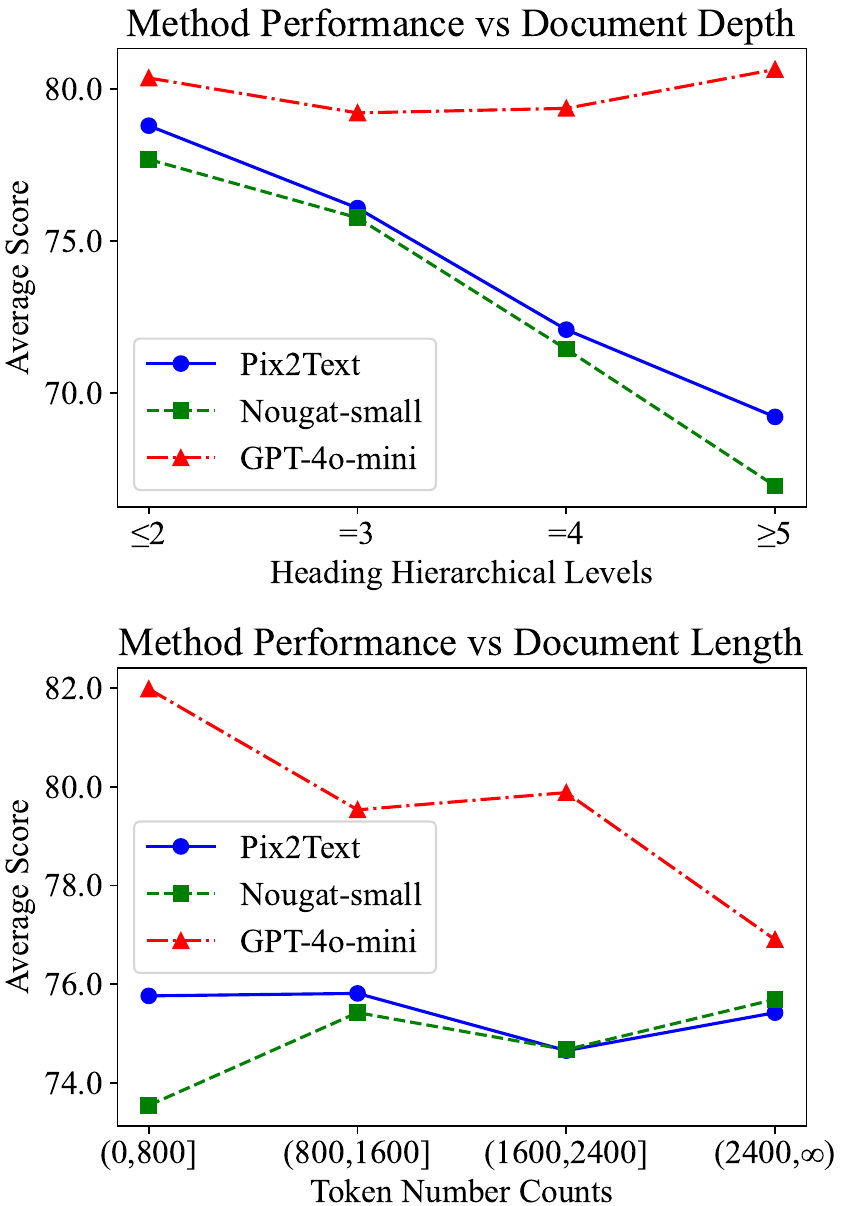}
 \caption{Relationship between DSE systems' performance and the depth or length of documents in \readoc-GitHub.}
 \label{fig.length_depth}
 \end{minipage}
\end{figure*}

\begin{table*}[ht] \small
  \centering
  \setlength{\tabcolsep}{1mm}
  \scalebox{0.95}{
    \begin{tabular}{c|cccccc|cc|c}
    \toprule
    \multirow{3}[2]{*}{\textbf{Methods}} & \multicolumn{6}{c|}{\textbf{Semantic Unit Evaluation}}             & \multicolumn{2}{c|}{\multirow{2}[2]{*}{\makecell{\textbf{Reading} \\ \textbf{Order}}}} & \multirow{3}[2]{*}{\textbf{Average}} \\ \cmidrule{2-7}
          & \multicolumn{2}{c}{\textbf{Text}} & \multicolumn{2}{c}{\textbf{Heading}}& \multicolumn{2}{c|}{\textbf{Table}} &  &  \\
          & \textit{Concat} & \textit{Vocab} & \textit{Concat} & \textit{Tree} & \textit{Concat} & \textit{Tree} & \textit{Block} & \textit{Token} &  \\
    \midrule
    MinerU & 57.28 & 59.95 & 30.73 & 22.83 & 38.75 & 26.82 & 65.46 & 66.87 & 46.08 \\
    Marker & 59.34  & 61.68  & 30.28  & 18.29  & 40.68  & 29.10  & 65.77  & 66.40  & 46.44 \\
    Nougat-base & 57.54 & 66.87 & \textbf{35.99} & \textbf{26.98} & 13.99  & 11.55 & 93.56  & 93.01  & 49.94 \\
    GPT-4o-mini & \textbf{64.16}  & \textbf{71.76}  & 25.07  & 15.4  & \textbf{45.75}  & \textbf{31.88}  & \textbf{95.23}  & \textbf{95.2}  & \textbf{55.56} \\
    \bottomrule
    \end{tabular} }
      \caption{Evaluation of three representative Document Structured Extraction systems on \readoc-Zenodo. }
  \label{tab:zenodo_result}%
\end{table*}%

\subsection{Compared Methods}

\paragraph{Baselines.}  We employ three simple and widely applicable methods as baselines. The first is PyMuPDF4LLM \cite{PyMuPDF4LLM}, a PDF bytecode parsing engine that converts digital-born PDFs into Markdown using embedded metadata. The second is Tesseract \cite{TessOverview}, an OCR tool for text extraction and basic page segmentation. The third is MarkItDown \cite{MarkItDown}, a versatile tool designed for converting various file types into Markdown format.

\paragraph{Pipeline Tools.} We evaluate three tools that support PDF-to-Markdown functionality, which integrate complex engineering with PDF parsing engines and advanced deep learning submodels. The pipeline tools we evaluate include the following. Marker \cite{marker}, MinerU \cite{2024mineru}, Pix2Text \cite{pix2text} and Docling \cite{auer2024doclingtechnicalreport}.


\paragraph{Expert Visual Models.} We evaluate Nougat-small and Nougat-base \cite{blecher2023nougat}, specialized transformer models trained on arXiv academic documents under the single-page image-to-Markdown paradigm, with parameter sizes of 250M and 350M, respectively. Additionally, we evaluate the GOT-OCR 2.0 model \cite{wei2024generalocrtheoryocr20}, a unified, end-to-end OCR system with 580M parameters.

\paragraph{Vision-Language Models.} We evaluate VLMs using the same single-page image-to-Markdown paradigm as expert models.
For efficiency, we select open-source models with fewer than 30 billion parameters and lightweight proprietary models.
Only VLMs with basic instruction-following and preliminary Markdown understanding are retained.
The retained VLMs include: the open-source models DeepSeek-VL-7B-Chat \cite{lu2024deepseek}, MiniCPM-Llama3-V2.5 \cite{viscpm}, LLaVa-1.6-Vicuna-13B \cite{liu2024llava}, InternVL-Chat-V1.5 \cite{chen2024far}, and the proprietary model GPT-4o-mini \cite{achiam2023gpt}.
More implementation details are included in Appendix \ref{exp_detail}.

\begin{table*}[ht] \small
  \centering
\setlength{\tabcolsep}{1mm}
\scalebox{0.96}{
    \begin{tabular}{c|c|cccccccc|cc|c}
    \toprule
    \multirow{3}[2]{*}{\textbf{Methods}} & \multirow{3}[2]{*}{\textbf{Modeling Paradigm}} & \multicolumn{8}{c|}{\textbf{Semantic Unit Evaluation}}          & \multicolumn{2}{c|}{\multirow{2}[2]{*}{\makecell{\textbf{Reading} \\ \textbf{Order}}}} & \multirow{3}[2]{*}{\textbf{Average}} \\
\cmidrule{3-10}          &       & \multicolumn{2}{c}{\textbf{Text}} & \multicolumn{2}{c}{\textbf{Heading}} & \multicolumn{2}{c}{\textbf{Formula}} & \multicolumn{2}{c|}{\textbf{Table}} & \multicolumn{2}{c|}{} &  \\
          &       & \textit{Concat} & \textit{Vocab} & \textit{Concat} & \textit{Tree} & \textit{Embed} & \textit{Isolate} & \textit{Concat} & \textit{Tree} & \textit{Block} & \textit{Token} &  \\
    \midrule
    \multirow{2}[1]{*}{GPT-4o-mini} & Single Page & \textbf{77.95} & \textbf{83.10} & 40.63 & 23.82 & \textbf{39.03} & \textbf{40.05} & \textbf{54.55} & \textbf{44.70} & 97.79 & \textbf{96.61} & 59.82  \\
    
          & Multiple Pages & 70.28 & 78.66 & \textbf{62.71} & \textbf{51.87} & 33.22 & 33.62 & 43.50  & 34.09 & \textbf{98.87} & 96.31 & \textbf{60.31} \\
    \bottomrule
    \end{tabular} }
      \caption{Comparison of page-level modeling paradigms, for documents within 5 pages of \readoc-arXiv.}
  \label{tab:multi_page}%
\end{table*}%

\subsection{Experimental Results}

Results for \readoc-arXiv, \readoc-GitHub and \readoc-Zenodo are presented in Table \ref{tab:arxiv_result} ,\ref{tab:github_result} and \ref{tab:zenodo_result}, respectively. We draw insights from both DSE system categories and specialized capabilities.

From the point of DSE system categories, we observe that: 
1) \textbf{Pipeline tools are often plagued by complex engineering challenges.} Docing falls short in recognizing embedded and isolated formulas, while Maker struggles with detecting embedded formulas, issues that are overlooked within the complex pipeline designs. In contrast, MinerU performs significantly better than other pipeline models in both embedded and isolated formula recognition. Besides, Pix2Text encounters program crashes and fails to process certain PDFs (3 files on \readoc-arXiv and 4 on \readoc-GitHub), posing a significant usability issue.
2) \textbf{Expert models struggle with generalization and scalability issues.} Nougat performs well on \readoc-arXiv but declines markedly on \readoc-GitHub (i.e., from 81.42 to 74.12) and \readoc-Zenodo (e.g., from 81.42 to 49.94), which features simpler layouts and fewer semantic units, indicating poor transfer learning ability. Moreover, scaling up from Nougat-small to Nougat-base does not boost performance (+0.04 on \readoc-arXiv, -0.74 on \readoc-GitHub). Additionally, while GOT-OCR 2.0 scores lower than Nougat-small and Nougat-base on \readoc-arXiv, it outperforms them on the \readoc-GitHub subset.
3) \textbf{VLMs generally underperform in complex academic documents.} The best-performing open-source model, InternVL-Chat-V1.5, scores 47.83 on \readoc-arXiv, while the proprietary model GPT-4o-mini scores 57.98, both of which are lower than the pipeline tools. On \readoc-Zenodo, GPT-4o-mini achieves an average score of 55.56, suggesting its potential in handling multilingual and multi-format documents.

From the perspective of specialized capabilities, we observe that: 
1) \textbf{Building hierarchical ToC trees from a global perspective remains a significant challenge,}
as existing systems predominantly focus on single-page images.
Pipeline tools lack modules to assess heading depth, leading to substantial drops in Tree EDS compared to Concat EDS. Expert models can exhibit strong ToC construction for specific documents, which is more a superficial imitation rather than a semantic understanding of the logical structure. On \readoc-arXiv, Nougat-base scores 88.50 in TEDS, while on \readoc-GitHub, it drops to 37.01.
2) \textbf{Understanding localized structured data such as tables and formulas is relatively difficult.} VLMs perform poorly on these tasks. Even the expert model Nougat-base, trained on arXiv documents, has shown only modest performance, with average metrics of 65.56 on these two tasks.
3) \textbf{Reading Order Detection is a relatively easy capability to acquire.} The baseline tool Tesseract, which uses heuristics for page segmentation, scores 96.70 and 98.48 in block-level KTDS on \readoc-arXiv and \readoc-GitHub, respectively.

\begin{figure*}[ht]
\centering
\includegraphics[height=2.3in]{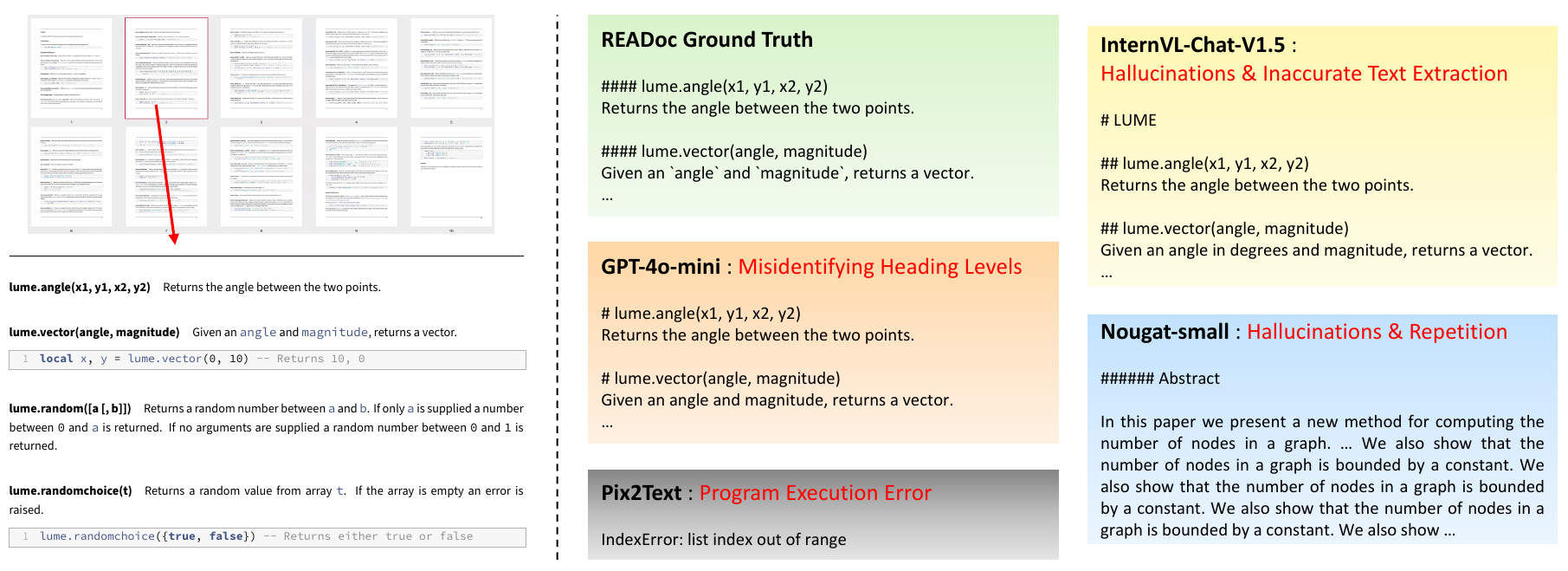}
\caption{A case study from READoc-GitHub. More cases are in Appendix \ref{exp_detail}.}
\label{fig.case}
\end{figure*}


\subsection{Fine-grained Analysis}

\paragraph{Impact of Document Length and Depth.}
Figure \ref{fig.length_depth} displays the results of three representative DSE systems on \readoc-GitHub. Pipeline tools and expert models exhibit similar performance trends, remaining stable with variations in document length but declining sharply as document depth increases. In contrast, VLMs demonstrate stability with changes in document depth, while their performance decreases as document length increases. This indicates that different DSE systems exhibit distinct shortcomings in realistic scenarios, which have not been previously revealed. 

\begin{table}[t] \small
  \centering
    \setlength{\tabcolsep}{1mm}
    \begin{tabular}{c|cc|c}
    \toprule
    \multirow{2}[3]{*}{\textbf{Methods}} & \multicolumn{3}{c}{\textbf{Semantic Unit Evaluation (Avg.)}} \\
\cmidrule{2-4}          & {Single-col.} & Multi-col. & Drop $\downarrow$ \\
    \midrule
    MinerU & 52.75  & 48.34 & 4.41 \\
    Pix2Text & \textbf{56.93} & \textbf{55.34} & \textbf{1.59} \\
    Marker & 56.70  & 53.39 & 3.31 \\
    \midrule
    Nougat-small & \textbf{80.51} & 74.09 & 6.42 \\
    Nougat-base & 80.48 & \textbf{74.16} & \textbf{6.32} \\
    \midrule
    InternVL-Chat-V1.5 & 39.96 & 32.83 & 7.13 \\
    GPT-4o-mini & \textbf{49.23} & \textbf{47.08} & \textbf{2.15} \\
    \bottomrule
    \end{tabular}%
      \caption{Relationship between DSE systems' semantic unit evaluation and the layout of documents in \readoc-arXiv.}
  \label{tab:column}%
\end{table}%

\begin{table}[t] \small
  \centering
    \setlength{\tabcolsep}{1mm}
    \begin{tabular}{c|c|cc}
    \toprule
    \multirow{2}[1]{*}{\textbf{Methods}} & \multirow{2}[1]{*}{\textbf{Modeling Paradigm}} & \multicolumn{2}{c}{\textbf{Heading}} \\
          &       & \textit{{Concat}} & \textit{{Tree}} \\
    \midrule
    Pix2Text & Single Page & 63.38 & 38.89 \\
    \midrule
    Nougat-small & Single Page & 64.87 & 39.53 \\
    \midrule
    \multirow{2}[1]{*}{GPT-4o-mini} & Single Page & 69.78 & 45.12 \\
          & Multiple Pages & \textbf{83.71} & \textbf{68.11} \\
    \bottomrule
    \end{tabular}%
      \caption{Comparison of page-level modeling paradigms, for documents within 5 pages of \readoc-GitHub.}
  \label{tab:github_pages}%
\end{table}%

\begin{table}[htbp] \small
  \centering
  \setlength{\tabcolsep}{1mm}
  \scalebox{0.93}{
    \begin{tabular}{c|c|c}
    \toprule
    \textbf{Methods} & \multicolumn{1}{l|}{ \makecell{\textbf{Time Cost} (s) \\ \textbf{per Document}}} & \makecell{\textbf{NVIDIA} \\ \textbf{GPU Devices}} \\
    \midrule
    Marker & 23.86 & \multicolumn{1}{c}{\multirow{5}[1]{*}{\makecell{{$1 \times$ Titan RTX} \\ {(24GB)}}}} \\
    MinerU & 30.96 & \\
    Nougat-small & 51.34 &  \\
    Nougat-base & 101.07 &  \\
    Pix2Text & 188.10 &  \\
    \midrule
    MiniCPM-Llama3-V2.5 & 392.37 & $1 \times$ A100 (80GB) \\
    InternVL-Chat-V1.5 & 1,182.02 & $2 \times$ A100 (80GB) \\
    \bottomrule
  \end{tabular} }
      \caption{Comparison of Efficiency of DSE systems.}
  \label{tab:efficiency}%
\end{table}%

\paragraph{Impact of Document Layout.} 
Using pdfplumber \cite{pdfplumber_2024} and heuristic rules, 
we classify documents in \readoc-arXiv into single- and multi-column categories, representing different layout complexities. Table \ref{tab:column} illustrates the average semantic unit evaluation scores across the two document types. All systems show performance degradation on complex multi-column documents, highlighting that our semantic unit evaluation implicitly measures the layout analysis capability. Among the systems, GPT-4o-mini exhibits the best layout analysis capability, while InternVL-Chat-V1.5 shows the most significant performance decline, reflecting substantial differences in performance levels among VLMs.

\paragraph{Exploration of the Multi-Page Paradigm.}
Previous researches focus on processing single pages and perform poorly in constructing global ToC trees. To explore how DSE systems might utilize global information, we investigate a paradigm where multiple pages are processed simultaneously. Specifically, we employ GPT-4o-mini to receive all page images of the document at once and convert them into Markdown text. We conduct experiments on documents with up to 5 pages, as shown in Tables \ref{tab:multi_page} and \ref{tab:github_pages}. While this method significantly enhances global ToC construction compared to the single-page paradigm, processing multiple pages simultaneously reduces local fine-grained capabilities, such as table and formula conversion, indicating that DSE modeling for multi-page documents still needs further development.

\paragraph{Considerations of Efficiency.}
DSE focuses on practical efficiency, which may involve real-time RAG calls or large-scale corpus construction. We sample 50 documents from \readoc-arXiv and measure the throughput of DSE systems, as shown in Table \ref{tab:efficiency}. Despite significant advancements in GPU memory and computational power, VLMs remains considerably lower efficiency compared to pipeline tools and expert Transformer models, indicating the need for future improvements in not only performance but also efficiency.

\subsection{Case Study}

We compare results from four representative systems in Figure \ref{fig.case}. Key observations include:
1) GPT-4o-mini misclassifies heading levels, revealing the limitations of single-page paradigms that fail to globally perceive the document's logical structure.
2) InternVL-Chat-V1.5 exhibits hallucinations and inaccurate text extraction, illustrating the differences in DSE capabilities between open-source and proprietary VLMs.
3) Nougat-small fabricates content completely unrelated to the original document, reflecting the poor generalization ability of expert models.
4) Pix2Text triggers an execution error, demonstrating the complexity in developing pipeline tools.

\section{Conclusion}

This paper proposes \readoc, a novel benchmark that frames document structured extraction as a realistic, end-to-end task, i.e., transforming unstructured PDFs into semantically rich Markdown text. 
Based on an evaluation S$^3$uite, We conduct a unified evaluation of state-of-the-art approaches, including pipeline tools, expert models and general VLMs.
Our experimental results reveal critical gaps in current researches when applied to realistic scenarios and underscore the importance of exploring new research paradigms.

\section*{Limitations}

Our work has two main limitations: 1) First, there exists some noise in the PDF-Markdown pairs generated through the automated framework. Although we have implemented various filtering and post-processing methods to minimize this impact, it remains difficult to eliminate completely. In fact, this is a common challenge in current document processing benchmark fields, and we will continue to explore more efficient and accurate processing paradigms in the future. 2) Second, although we introduce a Standardization module in our DSE evaluation S$^3$uite to unify the outputs from different models to the ground truth format, there are still some scenarios we cannot fully account for. We plan to introduce more comprehensive format unification modules in future work.

\section*{Ethics Statement}

All the data, tools and model weights we use come from publicly available sources. We only use them for evaluation purposes in document structured extraction. When using these resources for this study, we strictly adhere to their licensing agreements.

\section*{Acknowledgment}

We sincerely thank the reviewers for their insightful comments and valuable suggestions. This work was supported by Beijing Municipal Science and Technology Project (Nos. Z231100010323002), Beijing Natural Science Foundation (L243006), the Natural Science Foundation of China (No. 62306303, 62476265) and the Basic Research Program of ISCAS (Grant No. ISCAS-JCZD-202401).

\bibliography{custom}

\appendix





\section {Task Example} \label{task_example}

\begin{figure*}[t]
\centering
\includegraphics[width=0.94\textwidth]{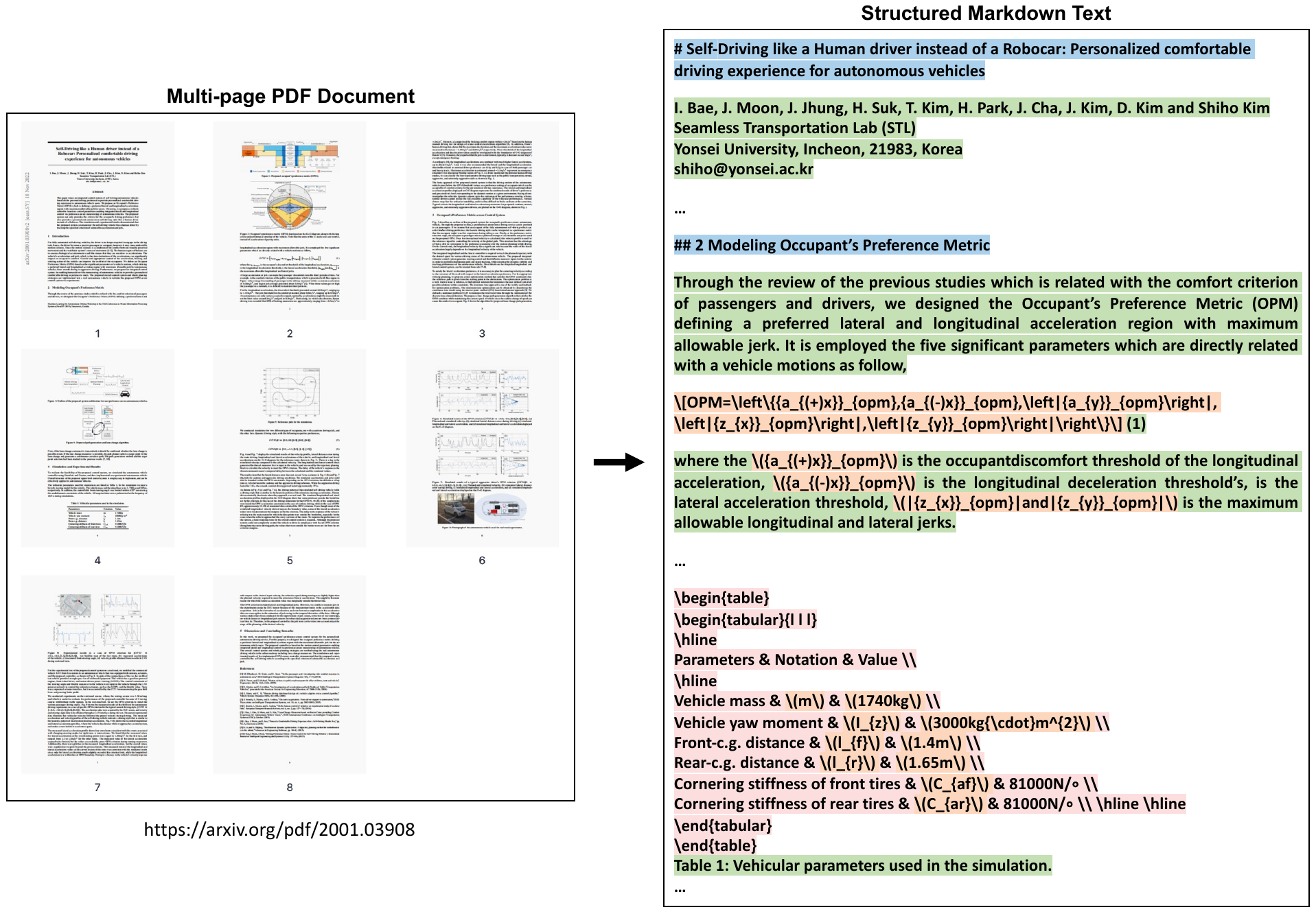} 
\caption{Input-Output example of \readoc~task.}
\label{fig.task}
\end{figure*}

This section provides an example of input and output for the Document Structured Extraction (DSE) task as defined by \readoc. As illustrated in Figure \ref{fig.task}, DSE systems are required to process a multi-page, real-world PDF document as input and produce a structured Markdown text as output.

\section{Details of Dataset} \label{dataset_detail}

\subsection{Construction Details}

\begin{figure}[ht]
\centering
\subfigure[]{
\centering
\includegraphics[height=2.1in]{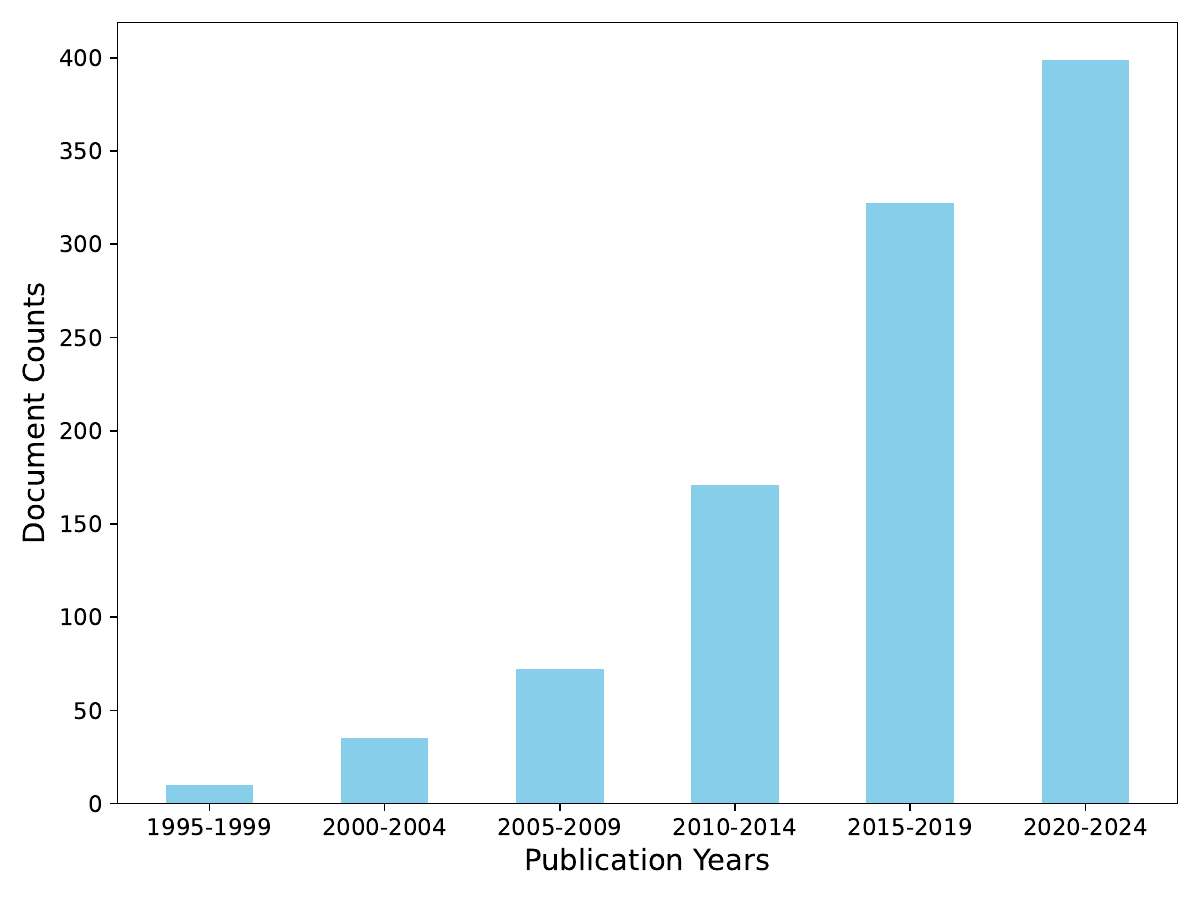}
}%

\subfigure[]{
\raggedright
\includegraphics[height=2.1in]{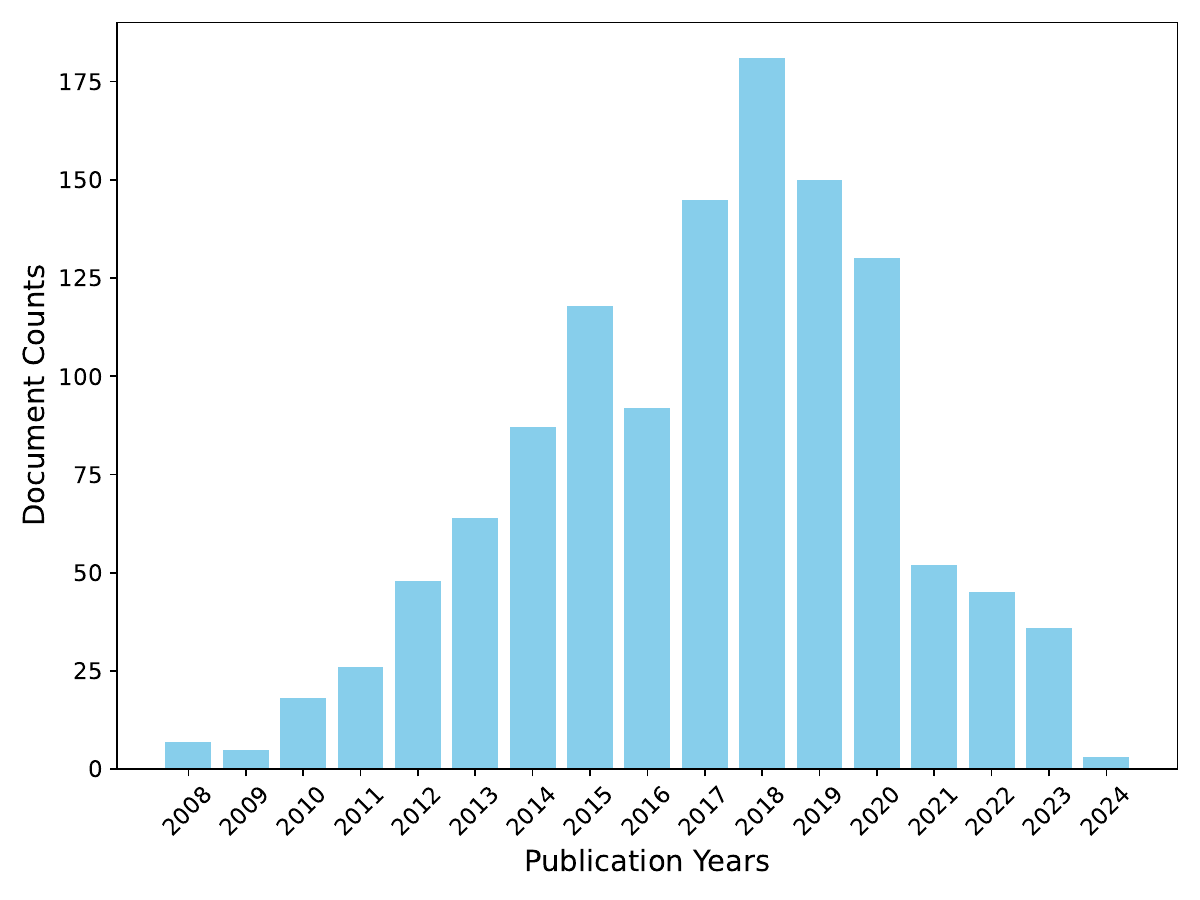}
}%
\centering
\caption{Visualization of year distribution in \readoc. (a) Document publication years of \readoc-arXiv. (b) Document publication years of \readoc-GitHub.}
\label{fig.year}
\end{figure}

\begin{figure*}[t]
\centering
\subfigure[]{
\centering
\includegraphics[height=2.in]{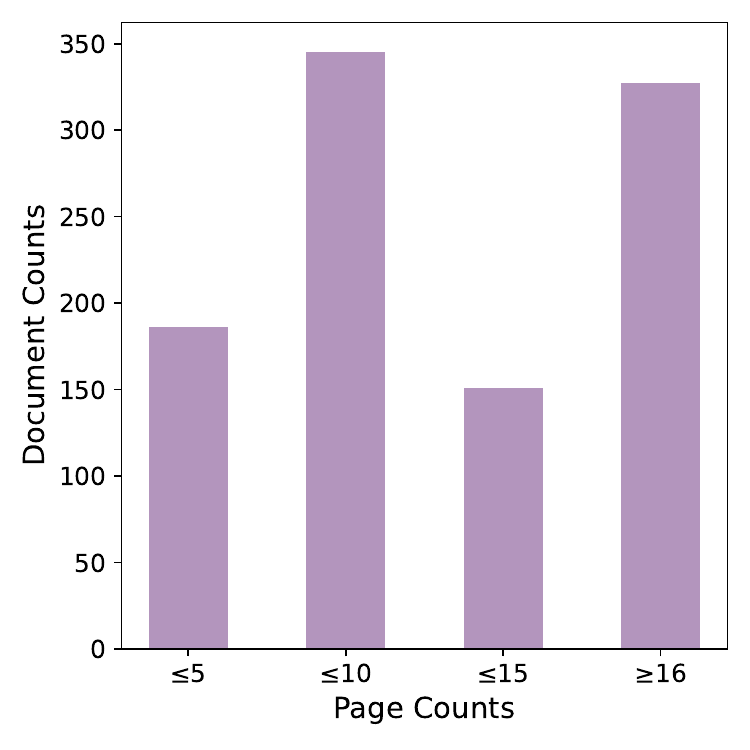}
}%
\subfigure[]{
\centering
\includegraphics[height=2.in]{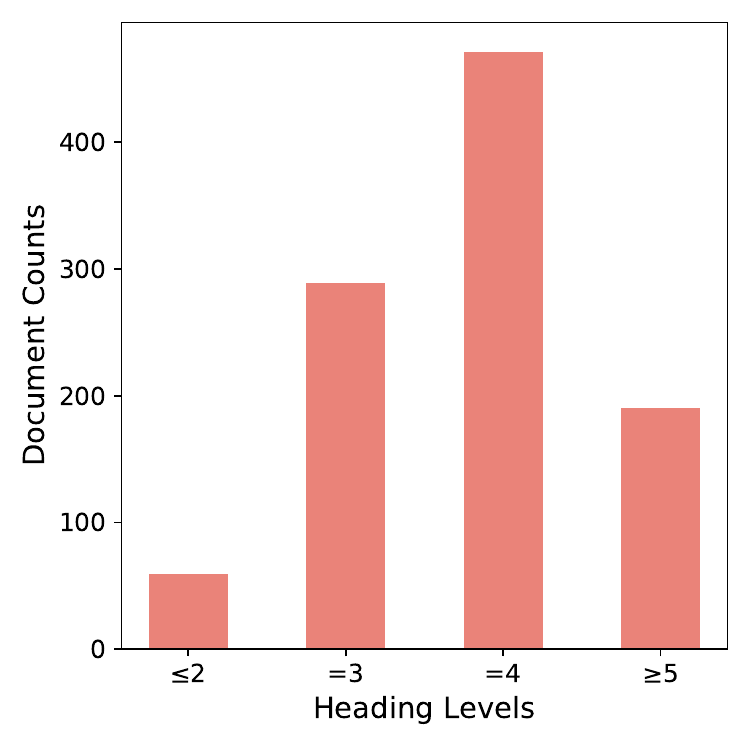}
}%
\subfigure[]{
\raggedright
\includegraphics[height=2.in]{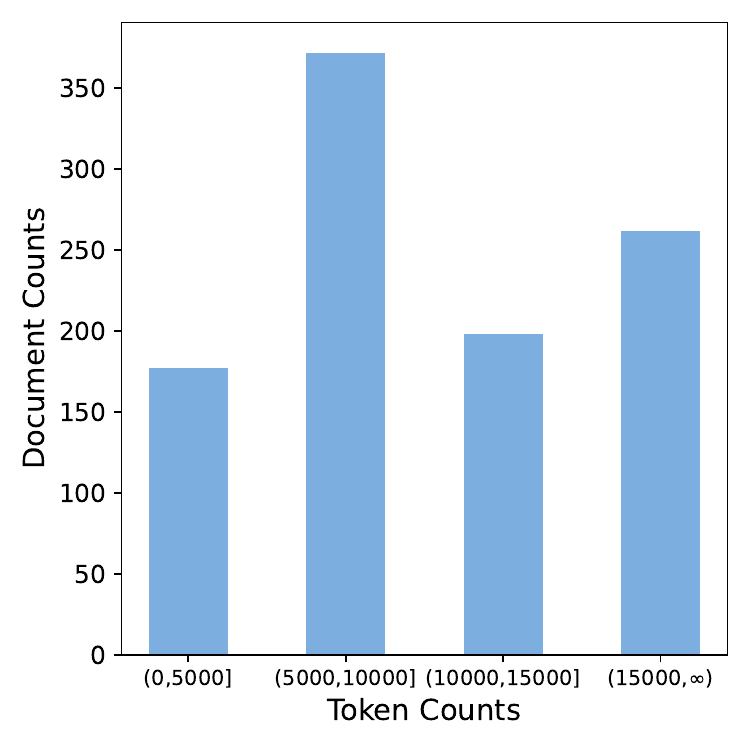}
}%
\centering
\caption{Visualization of year distribution in \readoc-arXiv. (a) Document page counts of \readoc-arXiv. (b) Document depth (heading levels) of \readoc-arXiv. (b) Document Length (token counts) of \readoc-arXiv.}
\label{fig.arxiv_count}
\end{figure*}

\begin{figure*}[t]
\centering
\subfigure[]{
\centering
\includegraphics[height=2.in]{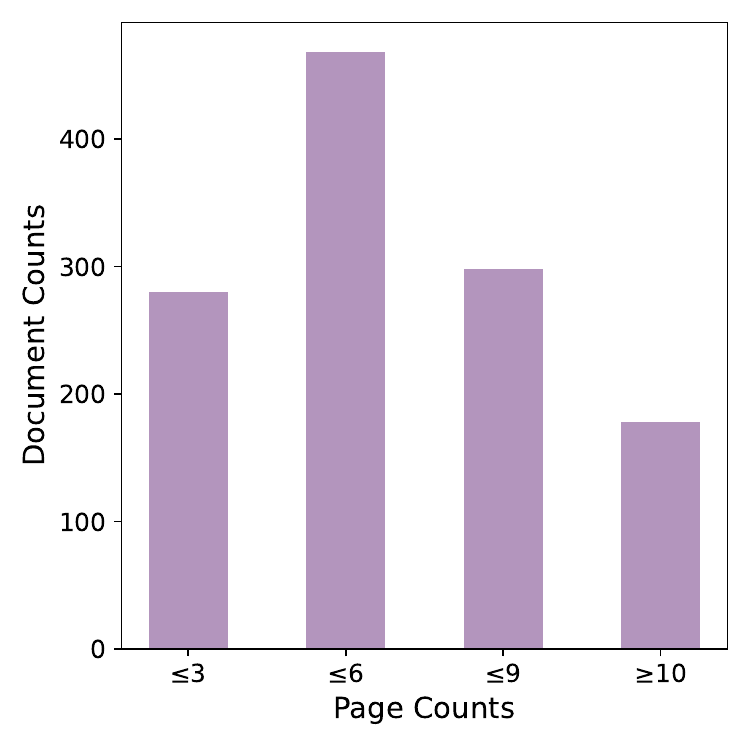}
}%
\subfigure[]{
\centering
\includegraphics[height=2.in]{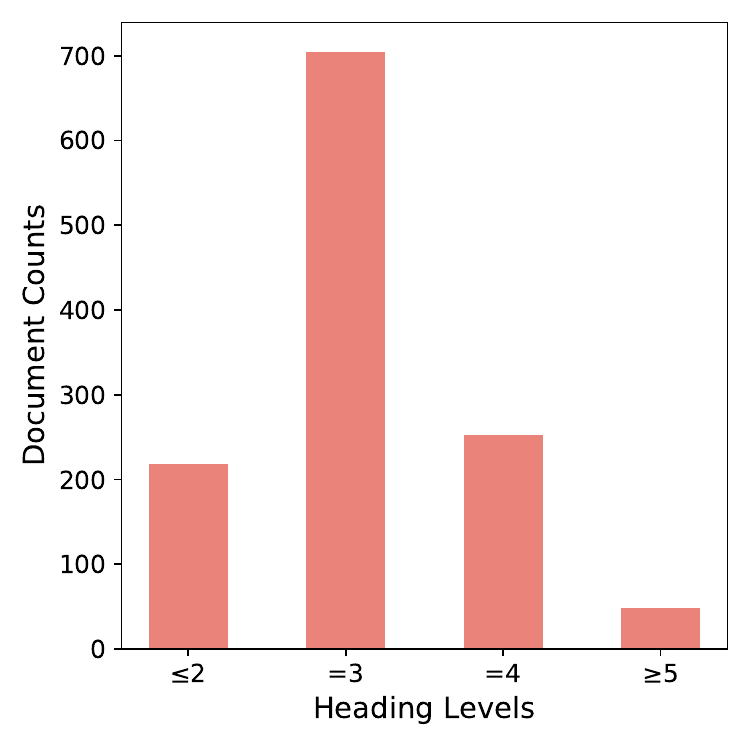}
}%
\subfigure[]{
\raggedright
\includegraphics[height=2.in]{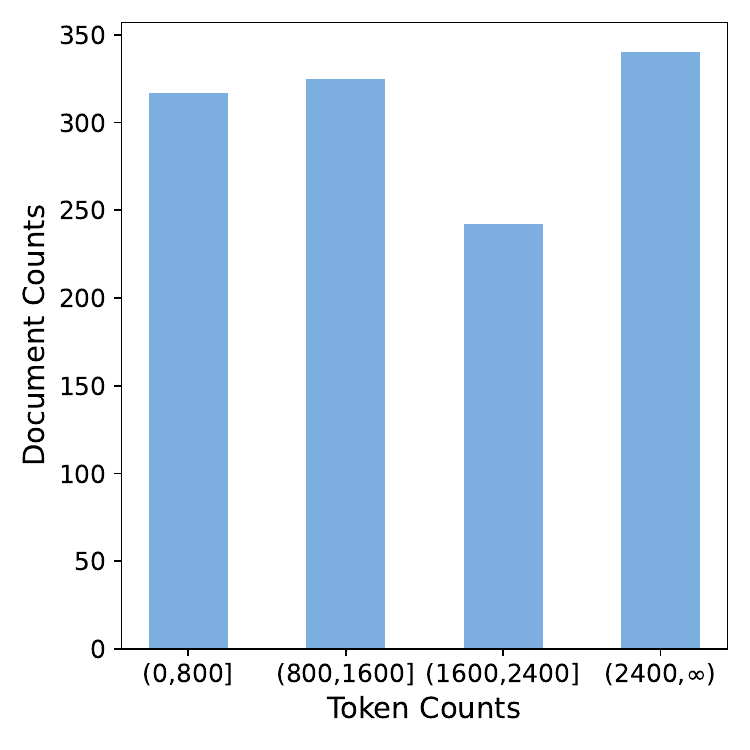}
}%
\centering
\caption{Visualization of year distribution in \readoc-GitHub. (a) Document page counts of \readoc-GitHub. (b) Document depth (heading levels) of \readoc-GitHub. (b) Document Length (token counts) of \readoc-GitHub.}
\label{fig.github_count}
\end{figure*}

\paragraph{Type Keywords of \readoc-arXiv.}

\begin{figure}[ht]
\centering
\includegraphics[height=2in]{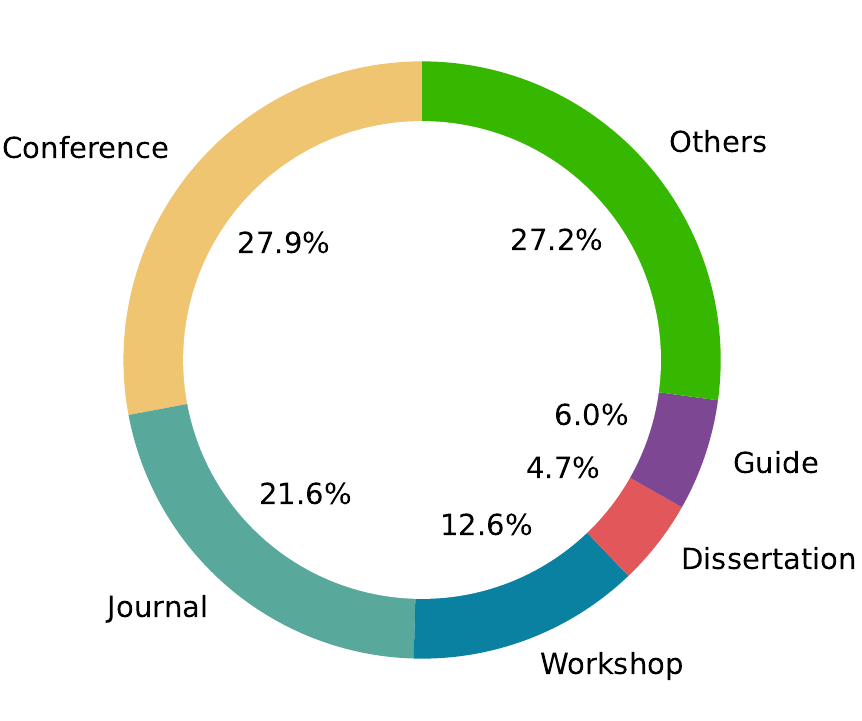}
\caption{Document types of \readoc-arXiv.}
\label{fig.arxiv_type}
\end{figure}

As we aim to improve data diversity in \readoc~across types, disciplines, topics, and so on, we use custom keywords to capture the document category types in \readoc-arXiv, as illustrated in Table \ref{tab:doc_classification}. We treat the comments of each arXiv preprint as retrieval targets, determining whether they contain the keywords to map them to specific document types.
The document type distribution of \readoc-arXiv is illustrated in Figure \ref{fig.arxiv_type}.

\begin{table}[h]
\centering
\begin{tabular}{c|c}
\toprule
\textbf{Document Type} & \textbf{Keywords} \\ \midrule
Workshop       & \textit{Workshop}     \\ \midrule
Conference     & \textit{Conference}   \\ \midrule
Journal        & \textit{Journal}      \\ \midrule
Dissertation   & \textit{Thesis, Dissertation}      \\ \midrule
Guide          & \makecell{\textit{Handbook, Manual,} \\ \textit{Guide, Tutorial,} \\ \textit{Technical Note}} \\ \midrule
Others         & -  \\ \bottomrule
\end{tabular}
\caption{Document type keywords of \readoc-arXiv.}
\label{tab:doc_classification}
\end{table}

\paragraph{Modification of Nougat process.}
As described in the main body of the paper, we modify Nougat’s source code \cite{blecher2023nougat} to convert arXiv documents in HTML format into Markdown text. Specifically, our modifications include: 1) improved support for converting the \textit{\textbackslash tableofcontents} command; 2) recognition and conversion of nested lists; 3) support for identifying and converting subtables and subfigures; and 4) removal of line breaks within headings in the Markdown format.

\subsection{Additional Data Statistics}

In addition to the diversity in disciplines, topics and languages, we provide supplementary statistical information here. This includes the distribution of the \readoc~dataset by year, as shown in Figure \ref{fig.year}, and the distribution by page count, depth, and length, as illustrated in Figure \ref{fig.arxiv_count} and Figure \ref{fig.github_count}.

\section{Details of Evaluation S$^3$uite} \label{suite_detail}

\subsection{Standardization Details}

We standardized the Markdown text output from various tools and models according to a set of specific rules:

\begin{itemize}
    \item \textbf{Alignment of different formula boundaries.} For isolated formulas, we standardize the starting boundary by converting \textit{\$\$}, \textit{\textbackslash begin\{equation\}}, \textit{\textbackslash begin\{gather\}}, 
    \textit{\textbackslash begin\{multline\}}, \textit{\textbackslash begin\{equation\}}, 
 to \textit{\textbackslash [}, with a similar format for the ending boundary. For embedded formulas, we convert the starting boundary \textit{\$} to \textit{\textbackslash (}, with a similar format for the ending boundary.
    \item \textbf{Heading formatting.} We standardize document headings by converting them to text lines beginning with consecutive ``\#" symbols, where the number of ``\#" indicates the heading level.
    \item \textbf{Removal of certain elements.} External URLs are removed from the link format and image references are also removed.
    \item \textbf{Table formatting.} We standardize tables by converting Markdown-compliant tables to LaTeX format. The reason for using LaTeX as the standardized table format is that, compared to native Markdown, it can represent more complex features such as multi-row and multi-column layouts.
\end{itemize}

\subsection{Segmentation Details}

\begin{figure}[t]
\centering
\includegraphics[width=0.35\textwidth]{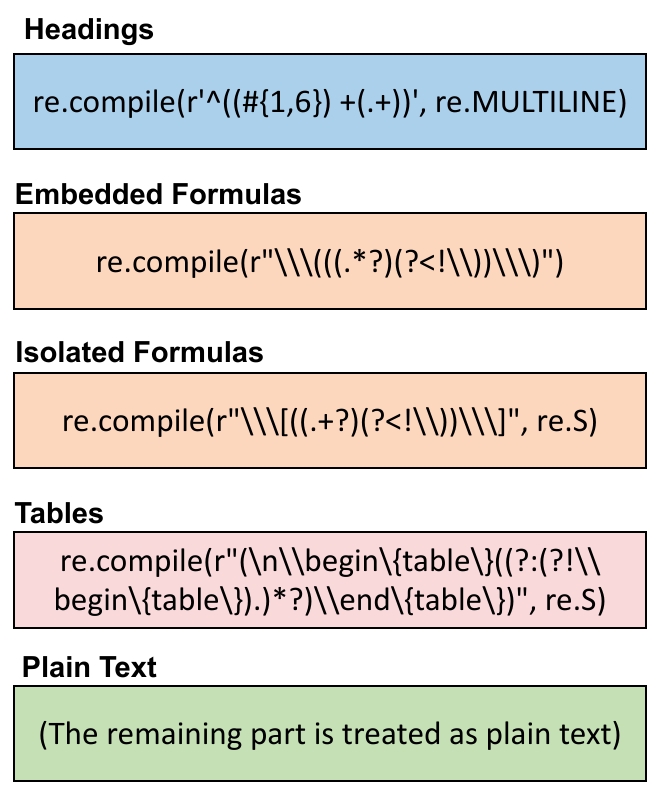} 
\caption{Segmentation module's regular expressions.}
\label{fig.seg}
\end{figure}

The segmentation module primarily based on a series of regular expressions written in Python, as illustrated in Figure \ref{fig.seg}.

\subsection{Scoring Details}

This section mainly discusses the metric calculations involved in the scoring module. Specifically, we focus on the following three similarity measures:

1) \textbf{Edit Distance Similarity (EDS)}: The edit distance \( ED(A, B) \) is defined as the minimum number of single-character edits (insertions, deletions, or substitutions) required to change string \( A \) into string \( B \). The edit distance similarity is calculated using the formula:

\[
EDS(A, B) = 1 - \frac{ED(A, B)}{\max(|A|, |B|)}
\]

where \( |A| \) and \( |B| \) are the lengths of strings \( A \) and \( B \), respectively.

2) \textbf{Tree Edit Distance Similarity (TEDS)}: The tree edit distance \( TED(T_1, T_2) \) is the minimum number of operations needed to transform one tree \( T_1 \) into another tree \( T_2 \). The operations typically include insertion, deletion, and relabeling of nodes. The tree edit distance similarity is computed as follows:

\[
TEDS(T_1, T_2) = 1 - \frac{TED(T_1, T_2)}{\max(|T_1|, |T_2|)}
\]

where \( |T_1| \) and \( |T_2| \) represent the sizes of trees \( T_1 \) and \( T_2 \), respectively.

3) \textbf{Kendall’s Tau Distance Similarity (KTDS)}: The Kendall’s Tau distance measures the ordinal association between two rankings. The number of discordant pairs \( K_d \) is defined as the number of pairs where \( x_i \) and \( x_j \) are in different orders. The Kendall’s Tau distance similarity is given by the formula:

\[
KTDS(X, Y) = 1 - \frac{2 \cdot K_d}{n(n-1)}
\]

where \( n \) is the total number of items.

\section{Details of Experiments} \label{exp_detail}

\subsection{Implementation of VLMs}

\begin{figure}[t]
\centering
\includegraphics[width=0.45\textwidth]{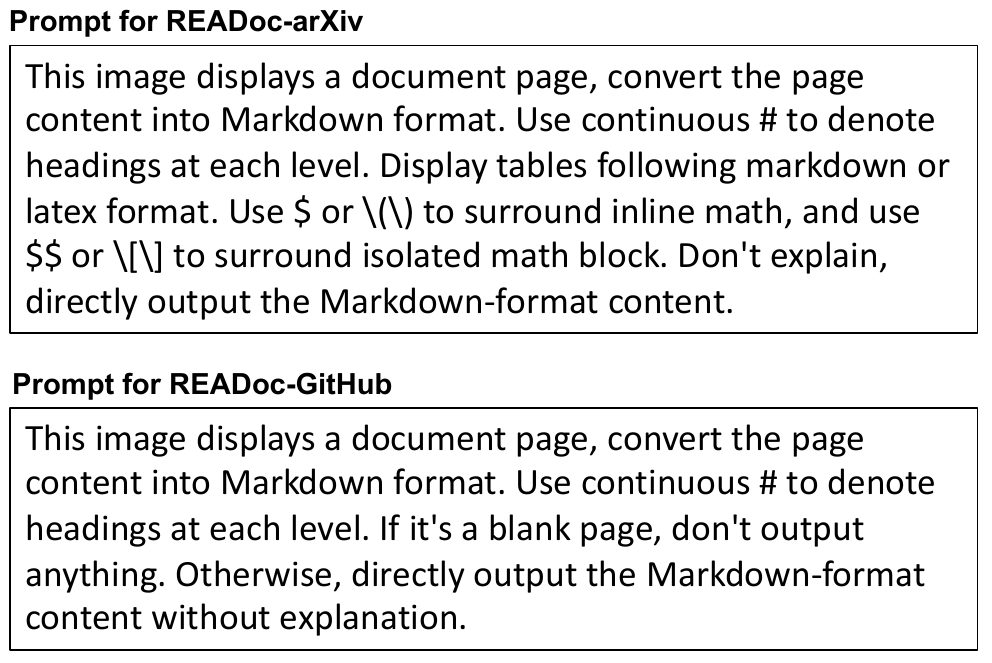} 
\caption{Prompts of Vision-Language Models.}
\label{fig.prompt}
\end{figure}

This section primarily supplements the evaluation process of Vision-Language Models. Specifically, we adopt a one-page image-to-Markdown approach, concatenating the Markdown text generated for each page. The prompts we use are illustrated in Figure \ref{fig.prompt}, and we uniformly apply the generation parameters for all VLMs as shown in Table \ref{tab:model_parameters}.

\begin{table}[h]
    \centering
    \begin{tabular}{c|c}
        \toprule
        \textbf{Parameter} & \textbf{Value} \\ \midrule
        top\_p & 0.8 \\ \midrule
        top\_k & 100 \\ \midrule
        temperature & 0.7 \\ \midrule
        do\_sample & True \\ \midrule
        repetition\_penalty & 1.05 \\ \bottomrule
    \end{tabular}
        \caption{Generation parameters of Vision-Language Models.}
    \label{tab:model_parameters}
\end{table}

\subsection{Additional Case Study}

\begin{figure*}[t]
\centering
\includegraphics[width=0.99\textwidth]{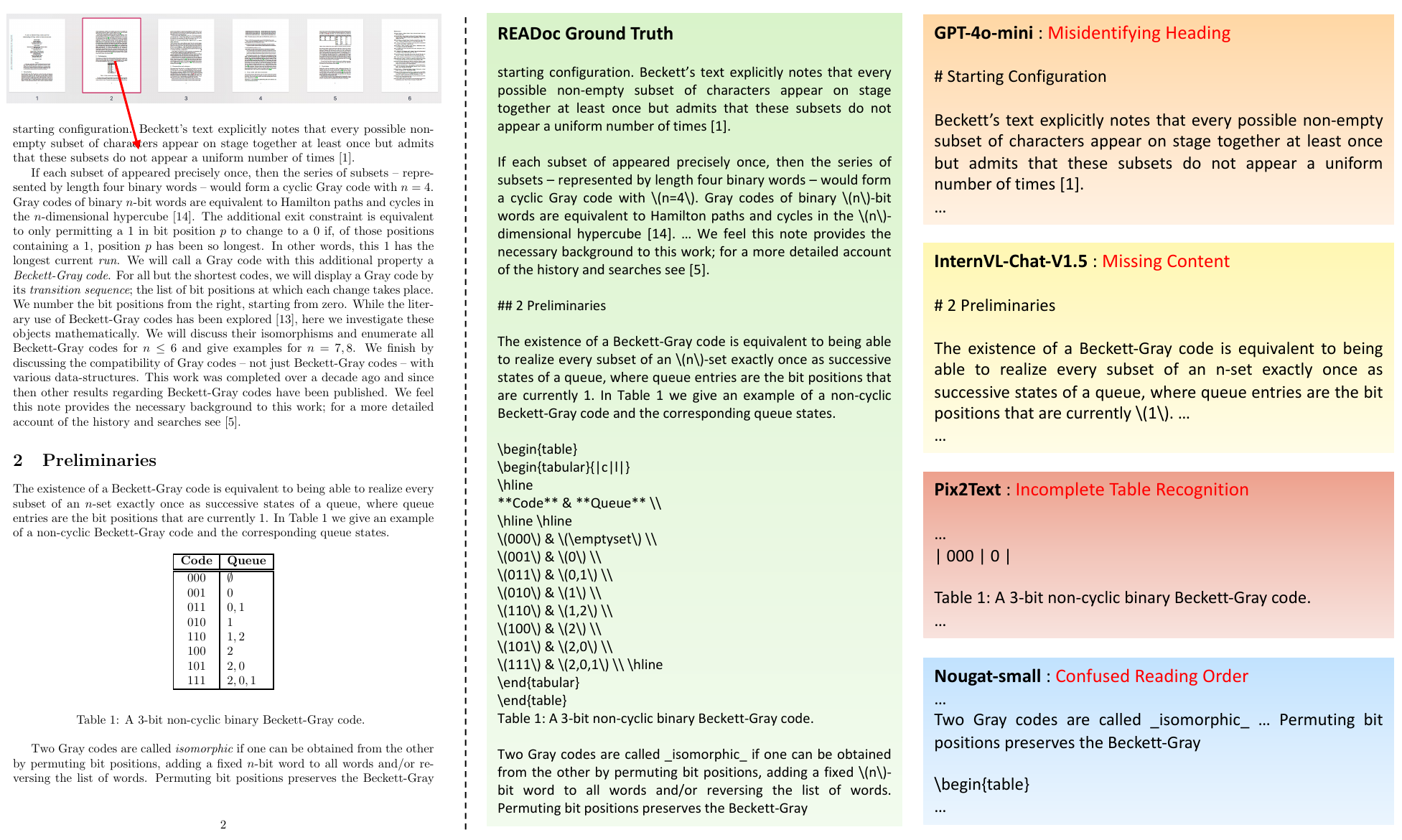} 
\caption{A case study from \readoc~arXiv.}
\label{case1}
\end{figure*}

\begin{figure*}[t]
\centering
\includegraphics[width=0.99\textwidth]{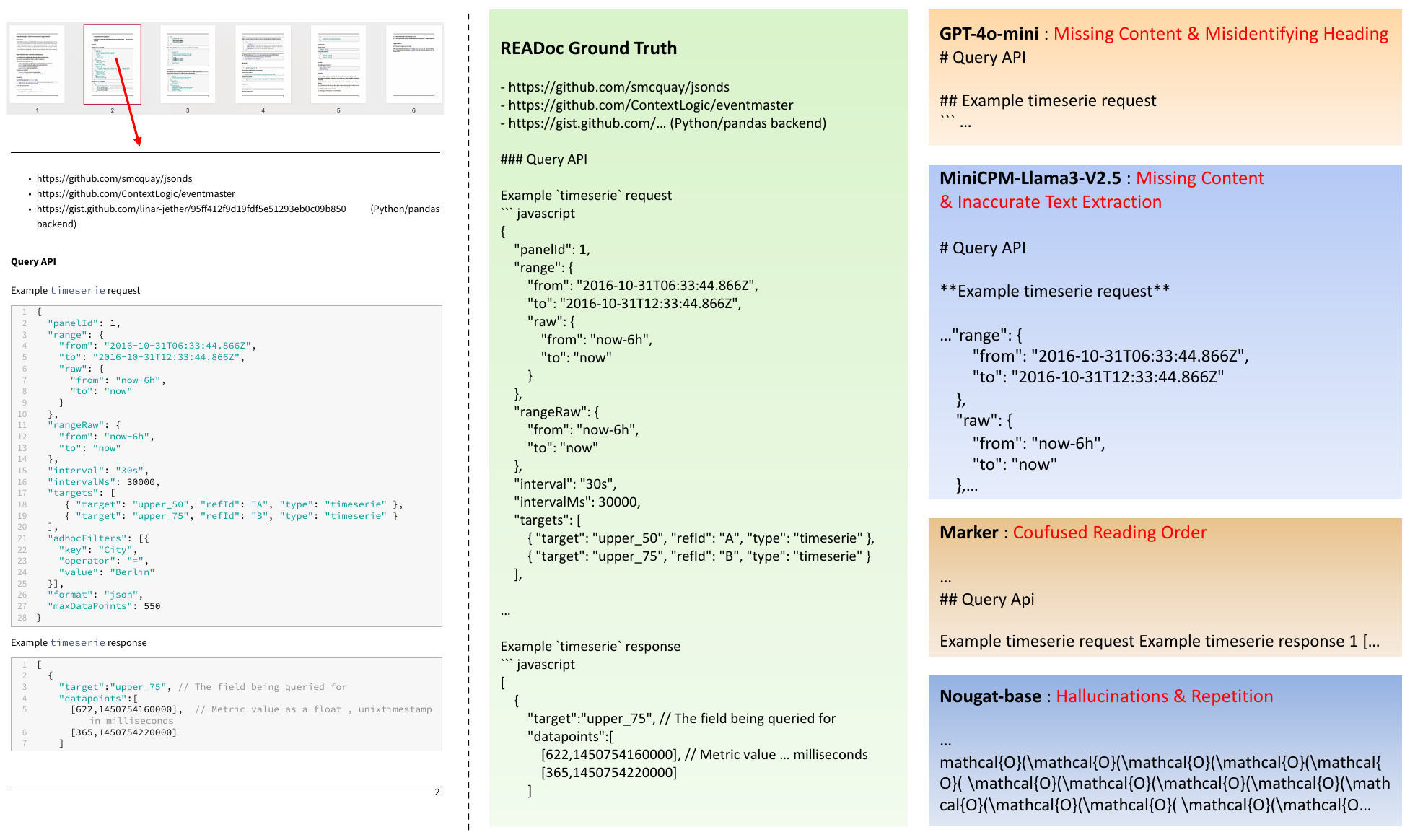} 
\caption{A case study from \readoc~GitHub.}
\label{case2}
\end{figure*}

We provide two additional case analyses, as shown in Figures \ref{case1} and \ref{case2}. It is evident that none of the DSE systems demonstrated satisfactory performance, as they exhibited various types of errors, including Missing Content, Misidentifying Headings, Inaccurate Text Extraction, Confused Reading Order, Hallucinations, and Repetitions.

\end{document}